\documentclass{article}

 \usepackage[preprint]{neurips_2026}

\usepackage[utf8]{inputenc} 
\usepackage[T1]{fontenc}    
\usepackage{hyperref}       
\usepackage{url}            
\usepackage{booktabs}       
\usepackage{amsfonts}    
\usepackage{amssymb} 
\usepackage{nicefrac}       
\usepackage{microtype}      
\usepackage{xcolor}         
\usepackage[table]{xcolor}

\usepackage{multirow}
\usepackage{graphicx}
\usepackage{makecell}

\usepackage{amsmath}
\usepackage{caption}
\usepackage{tikz}
\usepackage{pgfplots}
\usepgfplotslibrary{colormaps}
\pgfplotsset{compat=1.18} 
\usepackage{subcaption}
\usepackage{upgreek}
\usepackage{colortbl}
\usepackage{enumitem}

\usepackage{wrapfig}

\usepackage{pifont}

\newcommand{\unk}{\textcolor{gray}{\scriptsize unk}}

\usetikzlibrary{shapes.geometric}

\definecolor{when}{HTML}{15A802}

\definecolor{highcolor}{HTML}{9502A8}
\definecolor{lowcolor}{HTML}{15A802}

\definecolor{lwpurple}{HTML}{15A802}
\definecolor{tldrteal}{HTML}{a84202}
\definecolor{rltamber}{HTML}{9502a8}

\newcommand{\high}{\mathbf{\color{highcolor}{E}}}
\newcommand{\low}{{\mathbf{\color{lowcolor}S}}}

\usetikzlibrary{arrows.meta, positioning, calc, fit, backgrounds, patterns, decorations.pathreplacing, shadows.blur}

\title{LookWhen? Fast Video Recognition by\\Learning When, Where, and What to Compute}

\newcommand{\cofirst}{\textcolor{highcolor}{\scalebox{1.3}{$\bullet$}}}
\newcommand{\coadvise}{\textcolor{lowcolor}{\raisebox{1.3pt}{\scalebox{0.85}{$\bigstar$}}}}

\author{%
  \vspace{2pt}
  \textbf{\cofirst\kern1ptAli Salamatian$^1$ \quad \cofirst\coadvise\kern1ptAnthony Fuller$^{2,3}$} \\
  \vspace{2pt}
  \textbf{Pritam Sarkar$^{1,3}$ \quad James R. Green$^2$} \\
  \vspace{3pt}
  \textbf{\coadvise\kern1ptLeonid Sigal$^{1,3}$ \quad \coadvise\kern1ptEvan Shelhamer$^{1,2,3}$} \\
  \vspace{2pt}
  University of British Columbia$^1$ \quad Carleton University$^2$ \quad Vector Institute$^3$  \\
\cofirst\kern1pt Co-first author \quad \coadvise\kern1pt Co-advising author
}

\begin{document}

\maketitle

\vspace{-0.5cm} 
\begin{abstract}
Transformers dominate video recognition.
They split videos into tokens, and processing them has expensive superlinear computational cost.
Yet videos are filled with redundancy, so we can question the need for this expense.
We introduce LookWhen, a selector–extractor framework that factorizes video recognition into learning \emph{when}, \emph{where}, and \emph{what} to compute.
Our shallow selector gets a scaled-down video and quickly scores all tokens across space-time, while our deep extractor gets the top-K selected tokens to approximate full-video representations without actually processing all the tokens.
A key challenge is defining effective supervision for selection and extraction.
For selection pre-training, we introduce a score on representations that ranks tokens by uniqueness using a simple nearest-neighbor distance.
For extraction pre-training, we distill both a video teacher \emph{and} an image teacher, for which we normalize its frame-wise representations to learn \emph{what changes} within videos.
Through these strategies, our selector-extractor learns general and efficient representations for feature extraction or fine-tuning to a task.
Through experiments on Kinetics-400, SSv2, Epic-Kitchens, Diving48, Jester, and Charades, we show that LookWhen achieves a better accuracy-computation trade-off than efficient models and upgraded baselines of similar size.
LookWhen Pareto-dominates in accuracy-FLOPs on 9 of 12 cases (6 tasks $\times$ 2 settings) and roughly matches on 3.
In accuracy-throughput, measuring time in practice, LookWhen is more efficient still at 6.7$\times$ faster than InternVideo2-B at equal accuracy.\footnote{Code and pre-trained models: \url{https://github.com/alisalamatian1/LookWhen}}
\end{abstract}

\vspace{-0.2cm} 
\section{Introduction: Video computation takes too much time and space}
\vspace{-0.2cm} 

Transformers \cite{transformers} have revolutionized video modeling \cite{vivit, videomae, girdhar2021anticipative, videomaev2}.
They split videos into several thousand or more tokens for recognition.
Computational cost scales superlinearly with the number of tokens \cite{transformers}, which is a drawback for short videos and a real obstacle for long ones.
However, not all tokens are needed to compute accurate representations:
some tokens are redundant, some can be inferred by their surroundings, and others \emph{capture the scene}.
Transformers can choose among tokens, as they naturally handle sparse inputs, and their computation depends only on the number of tokens and not their distribution.
Sparsity can thus bring efficiency, but only if we can \textcolor{lowcolor}{\textbf{select}} \emph{when} and \emph{where} to process within each video so that we can \textcolor{highcolor}{\textbf{extract}} features accurately.

Factorizing \emph{when} (in time), \emph{where} (in space), and \emph{what} (in representation) is not easy.
It requires isolating the input tokens that are responsible for the output representation, and quickly, as otherwise it will not boost efficiency.
It also requires accurately representing the full video, given only the selected tokens, approximating dense computation with sparse computation.

We divide this approximation into token selection and feature extraction with a model for each.
Our efficient selector receives a \emph{downscaled} video, and scores all tokens in the full input.
Our expressive extractor receives \emph{only} the top-K \emph{selected} tokens, yet predicts features of the \emph{full} video.
Our \textbf{LookWhen} selector-extractor models learn representations of videos that neither model ever fully processes, surpassing the accuracy-compute trade-offs of existing video models.

The pre-training of a general selector-extractor was recently introduced for \emph{images} \cite{lookwhere}. 
Their models learn from a teacher: the selector is trained to predict the teacher's final attention map (\emph{where} to compute) and the extractor is trained to predict the teacher's final representations (\emph{what} to compute).
Video offers more redundancy than imagery, thus offering greater opportunity for efficiency via sparsity.
Directly learning from a teacher for \emph{video} recognition may not work: it needs a video teacher with relevant \cite{lookwhere} and \emph{artifact-free} \cite{registers} attention.
Attention to irrelevant tokens and artifacts is inefficient and ineffective for learning.

LookWhen is efficient due to how it selects, but selection is not new to video. 
Prior methods select to minimize \emph{pixel redundancy} \cite{rlt, k_center}.
We select to minimize redundancy by learning to \emph{predict token uniqueness}; where a token is unique if no other token in the video has similar features.
During pre-training, we measure uniqueness as the distance to a token's nearest neighbor in a \emph{teacher's feature space}. 
Our ``top1-distance'' method eliminates the need for artifact-free teacher attention and improves on existing and alternative targets for selection.

LookWhen is efficient because its predictions rely only on standard and highly-optimized operations.
Instead of selecting tokens at the input, other adaptive computation methods \emph{merge} tokens at intermediate layers \cite{tome, vid_tldr}. 
These methods are more reactive than predictive; they compute all tokens, match, then merge to continue with fewer.
They match by clustering and other operations, which can reduce FLOPs, but can fail to save time \cite{lookwhere}.

We make three main contributions:
\begin{itemize}[leftmargin=*, noitemsep, topsep=0pt]
\item We extend the selector–extractor framework to video with key changes to the architecture and pre-training.
We train to \textcolor{lowcolor}{\textbf{select}} unique tokens across space-time rather than the most attended tokens.
We train to \textcolor{highcolor}{\textbf{extract}} video-level tokens supervised by multiple teachers: a video teacher and an image teacher from which we make a target token through concatenating \emph{time-normalized} frame tokens to learn \emph{what changes} within each video for fine-grained classification.

\item We show LookWhen improves efficiency over existing video models and our own upgraded baselines.
Gains are highest for linear probing and are consistent across tasks: Kinetics-400 (K400) \cite{k400}, Something-Something-v2 (SSv2) \cite{ssv2}, Epic-Kitchens (EK100) \cite{ek-100}, Diving48 \cite{diving48}, Jester \cite{jester}, and Charades \cite{charades}.
LookWhen can even beat its non-sparse teacher: InternVideo2 (IV2) \cite{internvideo2}.

\item Through ablations on six datasets, we show that our novel methods for selector training (top1-distance) and extractor training (time-normalized frame features) drive LookWhen's performance.

\end{itemize}

\begin{figure}[t]
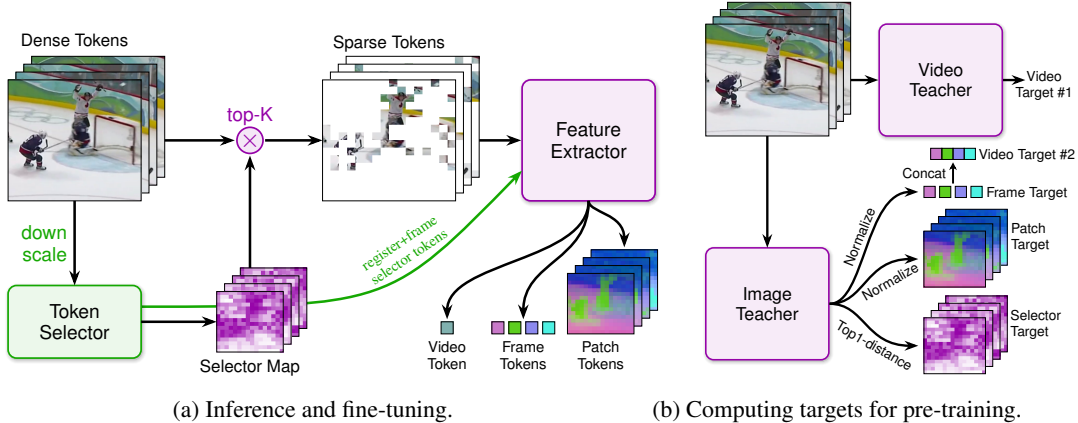

    \centering

    \begin{subfigure}[t]{0.58\textwidth}
        \centering
        \scalebox{0.8}{\input{figures/main/method.tex}}
        \caption{Inference and fine-tuning.}
        \label{fig:method}
    \end{subfigure}
    \begin{subfigure}[t]{0.4\textwidth}
        \centering
        \scalebox{0.74}{\input{figures/main/computing_targets}}
        \caption{Computing targets for pre-training.}
        \label{fig:training}
    \end{subfigure}

    \caption{%
\textbf{LookWhen's} \emph{shallow} \textcolor{lowcolor}{\textbf{selector}} gets a \emph{downscaled} video and scores tokens on their feature uniqueness (left).
Target uniqueness is from our ``top1-distance'' algorithm, which computes each patch's distance to its nearest neighbor in an image teacher's feature space (bottom right).
LookWhen's \textcolor{highcolor}{\textbf{extractor}} gets the top-K input tokens for \emph{sparse} and \emph{deep} processing.
Target features are from a video teacher (top right) and an image teacher (bottom right); we normalize to emphasize within-video \emph{change}.
Teachers are only needed during pre-training, so inference and fine-tuning is efficient.
    }
    \label{fig:overview}
\end{figure}

\section{LookWhen: Selecting across space \& time and extracting across teachers}

\subsection{Preliminaries: Look\emph{Where} and positionally-grounded representations}

\def\cell{0.105\textwidth}

\begin{figure}[t]
\centering
\setlength{\tabcolsep}{1.2pt}
\begin{tabular}{c c c c c c c c c}


\raisebox{16pt}{\makecell{Frames}}
& \includegraphics[width=\cell]{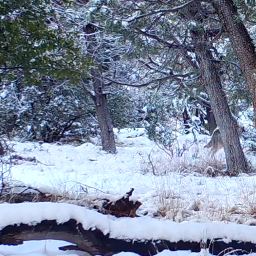}
& \includegraphics[width=\cell]{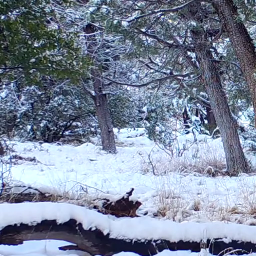}
& \includegraphics[width=\cell]{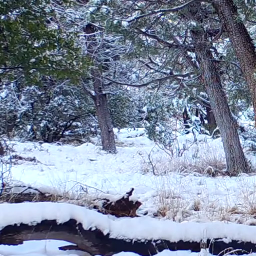}
& \includegraphics[width=\cell]{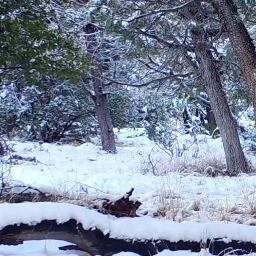}
& \includegraphics[width=\cell]{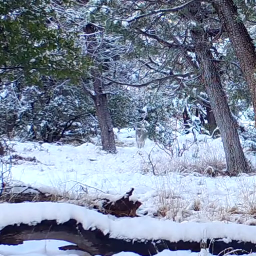}
& \includegraphics[width=\cell]{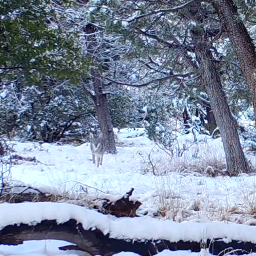}
& \includegraphics[width=\cell]{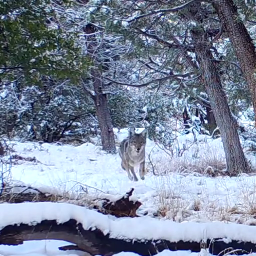}
& \includegraphics[width=\cell]{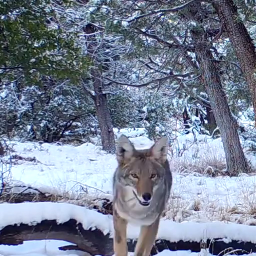} \\

\raisebox{16pt}{\makecell{Intern-\\Video2\\Attention}}
& \includegraphics[width=\cell]{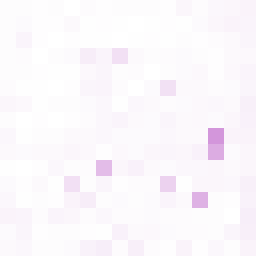}
& \includegraphics[width=\cell]{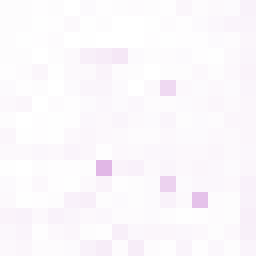}
& \includegraphics[width=\cell]{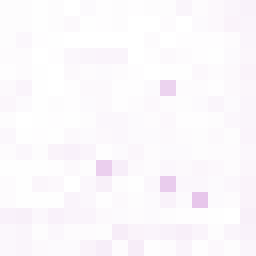}
& \includegraphics[width=\cell]{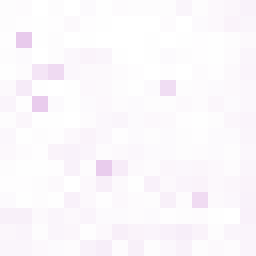}
& \includegraphics[width=\cell]{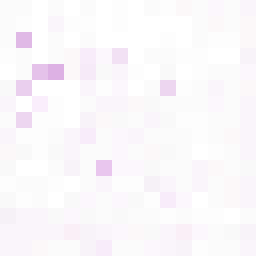}
& \includegraphics[width=\cell]{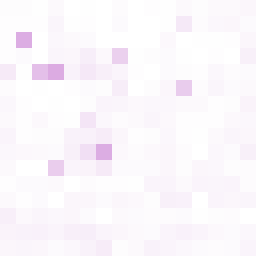}
& \includegraphics[width=\cell]{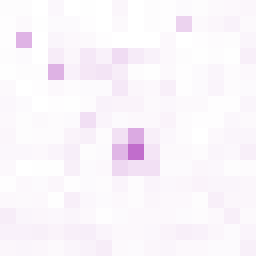}
& \includegraphics[width=\cell]{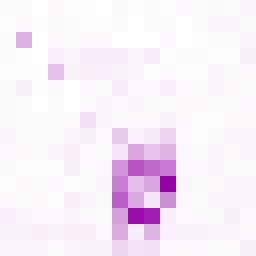} \\

\raisebox{16pt}{\makecell{DINOv3\\Attention}}
& \includegraphics[width=\cell]{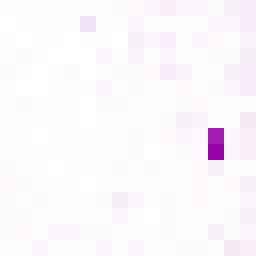}
& \includegraphics[width=\cell]{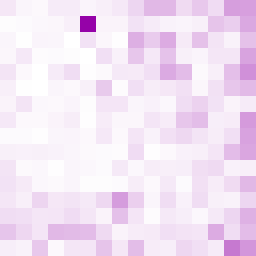}
& \includegraphics[width=\cell]{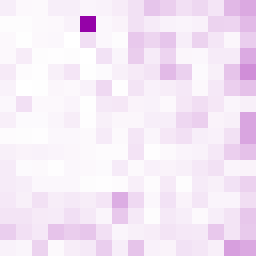}
& \includegraphics[width=\cell]{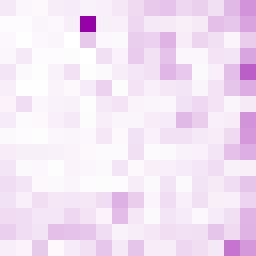}
& \includegraphics[width=\cell]{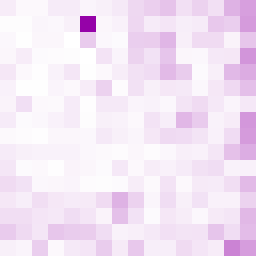}
& \includegraphics[width=\cell]{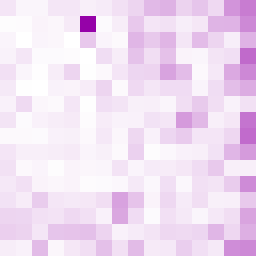}
& \includegraphics[width=\cell]{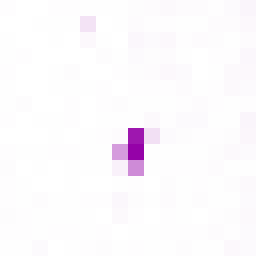}
& \includegraphics[width=\cell]{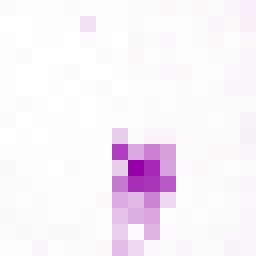} \\

\raisebox{16pt}{\makecell{DINOv3\\Top1-dist}}
& \includegraphics[width=\cell]{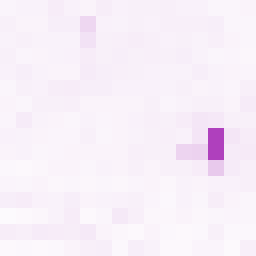}
& \includegraphics[width=\cell]{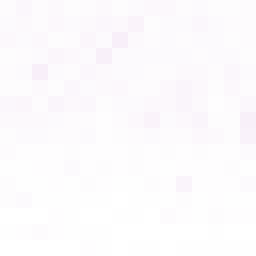}
& \includegraphics[width=\cell]{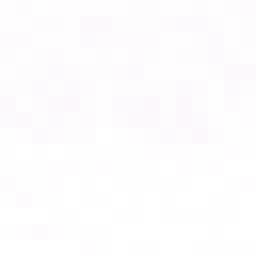}
& \includegraphics[width=\cell]{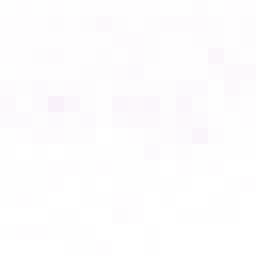}
& \includegraphics[width=\cell]{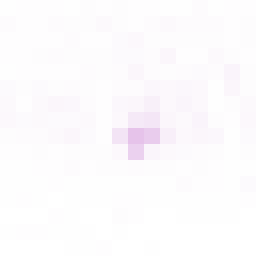}
& \includegraphics[width=\cell]{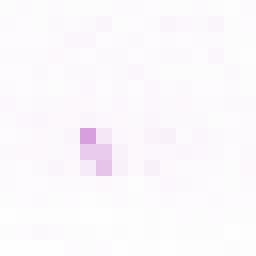}
& \includegraphics[width=\cell]{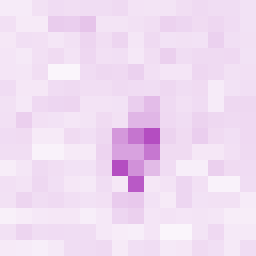}
& \includegraphics[width=\cell]{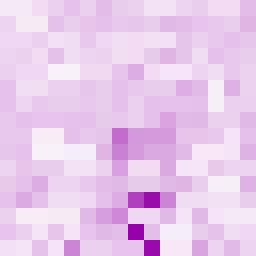} \\

\end{tabular}

\caption{
\textbf{When and where to compute for efficiency.}
InternVideo2 space-time attention maps suffer from \emph{artifacts}.
DINOv3 has cleaner attention but is strictly \emph{frame-wise}.
``Top1-dist'' (each patch's distance to its nearest neighbor in \emph{feature space}) finds the unique patches across all frames; our selector predicts it.
The wolf is partly visible in frame 1, runs away, then toward the camera.
}

\label{fig:story}

\end{figure}

\textbf{Look\emph{Where} inspires our Look\emph{When}.}
LookWhere \cite{lookwhere} introduced the selector-extractor framework.
The selector gets a downscaled image and outputs a selector map, for which the top-K positions are chosen.
The extractor gets the original-resolution image patches at the selected positions to extract features of the full image.
During pre-training, a teacher computes targets from \emph{all} original-resolution image patches.
The selector trains to predict the teacher's final attention maps.
The extractor trains to predict the teacher's final representations.
Thus they train in tandem to efficiently approximate the teacher's computation.
LookWhere outperforms prior adaptive computation methods at visual recognition; the gains are especially large for large images with more redundancy.
Videos have even more redundancy and are more expensive to process in full: this cries out for adaptive methods that choose where---and \emph{when}---to process.
However, to realize these potential efficiency gains, we need a teacher for which we can factorize representation (\emph{what}) and location (\emph{when} and \emph{where}).

\textbf{Positionally-grounded representations.}
Deep networks, including ViTs \cite{dino, tmlr_registers, registers, yang2024denoising, shi2026vision, yan2026vision}, can output feature maps that are positionally misaligned from the input.
That is, the feature vector at position $(x,y,t)$ may encode information that does not pertain to the input patch at $(x,y,t)$.
This misalignment manifests visibly: top-3 PCA projections of such feature maps appear noisy and unstructured.
``High-norm'' tokens \cite{registers} are the most well-known example, where certain patch tokens store global information unrelated to their position; understanding and eliminating high-norm tokens is an active area of research \cite{registers, jumbo}.
More broadly, optimization can redistribute features across local or non-local positions---e.g. shifting a patch's representation into a neighboring token---whenever doing so lowers the training loss.
This has grave consequences for designing selector-extractor methods, since their success relies on isolating the input tokens responsible for the output tokens. 

\subsection{Overview and motivation of the architecture and targets}

\textbf{Architecture (Fig. \ref{fig:overview}).}
The selector must see the full video to judge redundancy, but it must be fast.
So we make it \emph{shallow} and give it a \emph{downscaled} input, substantially reducing its cost.
The extractor must be sufficiently expressive to compute rich representations, so we give it full depth.
Its efficiency is from input sparsity: it receives only the top-K tokens that the selector deems most informative.
These two models approximate dense computation by exploiting the inherent redundancy in video.
We define three token types: a video token (1 per video), frame tokens (1 per frame), and patch tokens (1 per spatial position per frame).
We supervise them with video-level, frame-level, and patch-level features, respectively.
After pre-training, the teachers are discarded; only the efficient selector-extractor is needed.
The video token is the input to the linear head for downstream tasks.

\textbf{Selector training targets.}
Attention maps are a natural choice for selection targets, following LookWhere \cite{lookwhere}, but the attention of current video models is unsuitable.  
Fig. \ref{fig:story} shows InternVideo2 \cite{internvideo2} attention has artifacts.
Other potential video teachers are no better, e.g. V-JEPA-2 \cite{vjepa2} attention is noisy and VideoMAE \cite{videomae} attention does not outperform random selection \cite{lite}.
DINOv3 \cite{dinov3} has fewer attention artifacts but is an \emph{image} model and thus cannot directly exploit time redundancy.
Yet DINOv3 has spatially-grounded features: its output token at position $(x,y)$ represents the input at $(x,y)$ \cite{dinov3}. 
We achieve alignment in $(x,y,t)$ by processing each frame independently and stacking the resulting feature maps in time.
We can then rank tokens by how much unique information they carry: the extractor should get the least redundant tokens to best approximate dense representations.
We propose ``top1-distance'', which ranks each token by its distance to its nearest neighbor in feature space, so we train to select the most isolated tokens.
For example, Fig. \ref{fig:wes_selections} shows learned selection using this pre-training strategy.

\textbf{Extractor training targets.}
We want our extractor to learn general representations so it can be used as a feature extractor or efficiently fine-tuned on downstream tasks.
So we choose a SOTA video foundation model, InternVideo2, which is aligned with natural language embeddings.
We also choose a SOTA image foundation model, DINOv3, which is fully self-supervised.
Since we want our extractor to excel at fine-grained tasks, we normalize DINOv3 features so the targets represent \emph{what changes} throughout the video.
We normalize over space-time for patch targets and time for frames.

\subsection{Selector-Extractor: Pairing dense low-res computation with sparse high-res computation}\label{subsec:selector-extractor}

We shrink the \textcolor{lowcolor}{\textbf{selector's}} input and model depth for efficiency.
We downscale the video by 2$\times$ along its time, height, and width dimensions $T_\high{}{\times}R_\high{}{\times}R_\high{} \to T_\low{}{\times}R_\low{}{\times}R_\low{}$, resulting in 8$\times$ fewer patches.
We choose the 2$\times$ for simplicity.
We use only 3 transformer layers.
These two modifications reduce FLOPs by 50$\times$ at $T_\high{}$=16, $R_\high{}$=224.
The selector patchifies the video into a $T_\low{}{\times}N_\low{}{\times}N_\low{}$ grid, and prepends $T_\low{}$ frame tokens and $G$ registers.
The selector outputs latents $z_\low{} \in \mathbb{R}^{(T_\low{}+G+T_\low{}\cdot N_\low{}^2) \times D}$ and uses an MLP to compute the selector map $\hat{M} \in \mathbb{R}^{T_\high{}\times N_\high{}\times N_\high{}}$ from $T_\low{}\cdot N_\low{}^2$ patch tokens.

We shrink the \textcolor{highcolor}{\textbf{extractor's}} token count for efficiency.
The extractor receives \emph{only the top-K patches}---chosen by their selector-map scores---appended to the video and frame tokens.
Rather than using learned-embedding frame and register tokens as inputs, the extractor gets the \emph{selector}'s frame and register tokens, which transmits information between the two views.
Since the selector has $T_\low{}$ frame tokens but the extractor gets $T_\high{}$ frames, we upscale the selector's frame tokens via linear interpolation in time.
We also prepend a learned-embedding video token to the token sequence, which represents the full input.
The extractor processes these tokens and outputs features $z_\high{} \in \mathbb{R}^{(1 + T_\high{}+G+ \text{K}) \times D}$.

\def\cell{0.105\textwidth}

\begin{figure}[t]
\def\cell{0.105\textwidth}

\centering
\setlength{\tabcolsep}{1.5pt}
\begin{tabular}{c c c c c c c c c}

\raisebox{16pt}{\makecell{frames\\1-8}}
& \includegraphics[width=\cell]{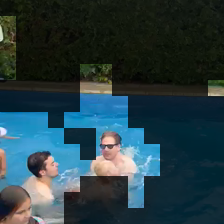}
& \includegraphics[width=\cell]{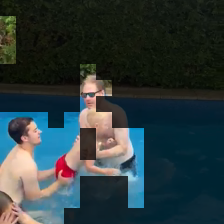}
& \includegraphics[width=\cell]{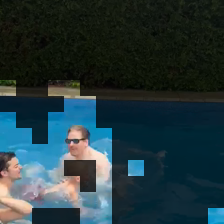}
& \includegraphics[width=\cell]{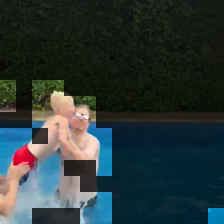}
& \includegraphics[width=\cell]{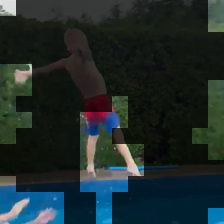}
& \includegraphics[width=\cell]{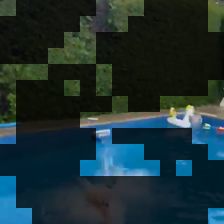}
& \includegraphics[width=\cell]{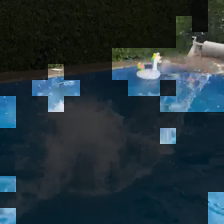}
& \includegraphics[width=\cell]{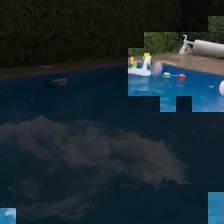} \\

\raisebox{16pt}{\makecell{frames\\9-16}}
& \includegraphics[width=\cell]{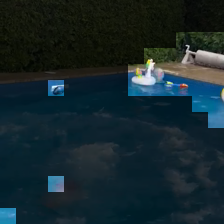}
& \includegraphics[width=\cell]{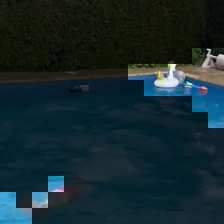}
& \includegraphics[width=\cell]{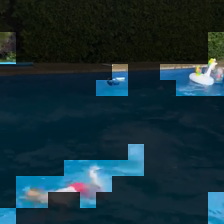}
& \includegraphics[width=\cell]{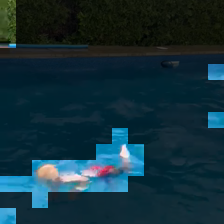}
& \includegraphics[width=\cell]{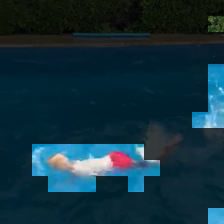}
& \includegraphics[width=\cell]{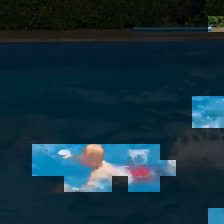}
& \includegraphics[width=\cell]{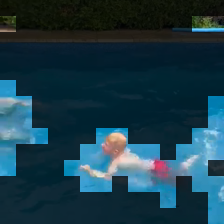}
& \includegraphics[width=\cell]{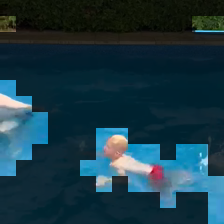} \\

\end{tabular}
\caption{
\textbf{Example of learned selections.}
We pre-train on K400+SSv2 data and our selector generalizes to a video of an author's nephew being thrown in a pool and swimming.
More in \S \ref{app:selection_examples}.
}

\label{fig:wes_selections}

\end{figure}

\subsection{Teachers: Pre-training to select unique patches and extract dense features}\label{subsec:teachers}

We train the \textcolor{lowcolor}{\textbf{selector}} to pick unique patches to drop redundant information in the video for efficiency.
To find the unique patches, we first compute features for all $T_\high{}{\cdot}N_\high{}^2$ patches in a video.
We extract them using an \emph{image} teacher, DINOv3 \cite{dinov3}, which has spatially-grounded representations.
We then apply our ``top1-distance'' algorithm, which computes each patch's cosine similarity to its nearest neighbor---in feature space---then orders all patches by increasing similarity so the most isolated patch is ranked first.
Formally, $U_{x,y,t}{=}1{-}\max_{(x',y',t') \neq (x,y,t)} \cos\!\left(z^{\text{DINOv3}}_{x,y,t},\, z^{\text{DINOv3}}_{x',y',t'}\right)$, where $U$ is the uniqueness score and $z^{\text{DINOv3}}_{x,y,t}$ is DINOv3's feature vector at patch position $(x,y,t)$.
We then post-process these uniqueness scores so we can train with binary cross-entropy (BCE) loss; specifically, we assign the lowest-ranking patch a 0, the highest a 1, and evenly space scores in between.
These steps result in a selector-map target $M \in \mathbb{R}^{T_\high{}\times N_\high{}\times N_\high{}}$, and we compute the loss $\mathcal{L}_{\text{map}}{=}\text{BCE}(\hat{M}, M)$.

We train the \textcolor{highcolor}{\textbf{extractor}} to represent the full video despite only processing some of it for efficiency.
We compute video features from two frozen teachers: a video teacher, InternVideo2, and an image teacher, DINOv3.
We choose InternVideo2's attention-pooled token $z^{\text{IV2}}_{\text{video}} \in \mathbb{R}^{D}$.
Since DINOv3 is an image model, it does not have a video-level token. 
To make one, we first normalize DINOv3's $T_\high{}$ class tokens over time so each dimension has 0 mean and 1 standard deviation.
We then concatenate these normalized class tokens, resulting in the other video-token target $z^{\text{DINOv3}}_{\text{video}} \in \mathbb{R}^{T_\high{}\cdot D}$.
We map from the extractor's video token to these targets with two separate MLPs and compute the mean-squared error (MSE) loss $\mathcal{L}_{\text{video}}
{=}\frac{1}{2}(\text{MSE}(\hat{z}^{\text{IV2}}_{\text{video}}, z^{\text{IV2}}_{\text{video}})
{+}\text{MSE}(\hat{z}^{\text{DINOv3}}_{\text{video}}, z^{\text{DINOv3}}_{\text{video}}))$.

We also train the \textcolor{highcolor}{\textbf{extractor's}} \emph{patch} and \emph{frame} tokens so we can approximate all teacher outputs.
For frame-token targets, we use DINOv3's class tokens, normalized over time.
For patch-token targets, we use DINOv3's patch tokens, normalized over space and time.
Since the extractor processes K patches, we upsample with nearest neighbor to make the full feature map $\hat{z}^{\text{DINOv3}}_{\text{patch}} \in \mathbb{R}^{T_\high{}\times N_\high{}\times N_\high{} \times D}$, following LookWhere.
Normalizing DINOv3's features emphasizes feature-change \emph{within} each video.
We also compute these losses $\mathcal{L}_{\text{frame}}{=}\text{MSE}(\hat{z}^{\text{DINOv3}}_{\text{frame}}, z^{\text{DINOv3}}_{\text{frame}})$ and $\mathcal{L}_{\text{patch}}{=}\text{MSE}(\hat{z}^{\text{DINOv3}}_{\text{patch}}, z^{\text{DINOv3}}_{\text{patch}})$.

\textbf{Selection and extraction losses.}
We train the selector and extractor simultaneously to minimize $\mathcal{L}{=}
\mathcal{L}_{\text{map}}{+}\mathcal{L}_{\text{video}}{+}\mathcal{L}_{\text{frame}}{+}\mathcal{L}_{\text{patch}}$.
Gradients flow from the extractor to the selector only through the selector's frame and register tokens, which are passed between the models.
Gradients do not flow from the extractor to the \emph{selector map}, since top-K patch selection is non-differentiable. 
We run DINOv3 online, which requires one forward pass per frame to compute all targets for selection and extraction.
We pre-compute InternVideo2's video tokens over all pre-training data to save time.

\section{Experiments: Accuracy and efficiency in FLOPs and time}
\label{experiments}

\textbf{Pre-training.}
We pre-train LookWhen for 20 epochs on the combined Kinetics-400 \cite{k400} and SSv2 \cite{ssv2} datasets.
We initialize our selector-extractor from DINOv3, and encode time with 1D-sincos embeddings.
Since training is bottlenecked by both data loading and online teacher processing, we repeat batches 8$\times$ via data augmentation so our selector-extractor sees more data (repeat augmentation is commonly used when training video models \cite{videomaev2, MaskedAutoencodersSpatiotemporal2022}).
For each batch we uniformly sample the sparsity level $S\in[70\%,95\%]$ so LookWhen supports varied sparsity out of the box; $S$\% sparsity means $\text{K}{=}(1{-}S){\cdot}N_\high{}^2$.
We choose a ViT-B size for our selector-extractor and teachers for feasibility.
We use $T_\high{}$=16 timesteps and $R_\high{}$=224 image size because that is most common in the literature.

\subsection{Downstream Tasks: Accuracy versus computational cost}

\begin{figure*}[t!]
\centering
\begin{minipage}[c]{0.62\textwidth}
\centering
\scriptsize
\setlength{\tabcolsep}{1pt}
\begin{tabular}{lcrrrr}
&
& \multicolumn{2}{c}{Kinetics-400}
& \multicolumn{2}{c}{SSv2}
\\
\cmidrule(lr){3-4}
\cmidrule(lr){5-6}
& Params
& FLOPs
& Top-1
& FLOPs
& Top-1
\\
Model
& \scriptsize{M}
& \scriptsize{G$\times$T$\times$S}
& \scriptsize{\% acc.}
& \scriptsize{G$\times$T$\times$S}
& \scriptsize{\% acc.}
\\
\midrule
\multicolumn{6}{l}{\emph{Larger models for reference only, not for direct comparisons}}\\
V-JEPA-2 \cite{vjepa2} & 355 & 935$\times$8$\times$3 & 85.1 & 935$\times$2$\times$3 & 73.7 \\
InternVideo2 \cite{internvideo2} & 1020 & 2500$\times$4$\times$3 & 89.4 & 2500$\times$2$\times$3 & 69.7 \\
\arrayrulecolor{gray!50}\midrule\arrayrulecolor{black}
\multicolumn{6}{l}{\emph{ViT-B, Swin-B, or Mamba-M models}}\\
UMT-B800e \cite{umt} & 87 & 180$\times$4$\times$3 & 85.7 & 180$\times$2$\times$3 & 70.8 \\
VideoMAE \cite{videomae} & 87 & 180$\times$5$\times$3 & 81.5 & 180$\times$2$\times$3 & 70.8 \\
VideoMAEv2 \cite{videomaev2} & 87 & 180$\times$5$\times$3 & 81.5 & 180$\times$2$\times$3 & 71.2 \\
VideoMamba-M800e \cite{videomamba} & 74 & 202$\times$4$\times$3 & 83.4 & 202$\times$2$\times$3 & 71.0 \\
VideoMambaPro \cite{lu2025snakes} & 72 & 392$\times$4$\times$3 & 84.0 & 183$\times$4$\times$3 & 69.4 \\
VideoSwin + STTS ($T_0^{0.6}$) \cite{stts} & 89 & 181$\times$4$\times$3 & 81.4 & 190$\times$1$\times$3 & 68.1 \\
VideoMAE + LITE (K=0.3) \cite{lite} & 87 & 46$\times$5$\times$3 & 78.4 & 46$\times$2$\times$3 & 68.3 \\
VideoMAE + ToMe (r=64) \cite{tome, rlt} & 87 & 131$\times$4$\times$3 & 80.0 & 131$\times$4$\times$3 & 69.7 \\
VideoMAE + RLT ($\tau$=0.1) \cite{rlt} & 87 & 120$\times$4$\times$3 & 80.1 & 120$\times$4$\times$3 & 70.2 \\
VideoMAE + vid-TLDR \cite{vid_tldr} & 87 & --- & --- & 57$\times$\unk & 69.6 \\
\rowcolor{when!20} \textbf{LookWhen} (90\% sparse) & 106 & 40$\times$4$\times$3 & 82.6 & 40$\times$2$\times$3 & 69.3 \\
\rowcolor{when!20} \textbf{LookWhen} (70\% sparse) & 106 & 108$\times$4$\times$3 & 84.6 & 108$\times$2$\times$3 & 72.0 \\
\end{tabular}
\end{minipage}%
\hfill
\begin{minipage}[c]{0.35\textwidth}
\captionof{table}{%
\textbf{LookWhen achieves a better inference accuracy-computation trade-off for fine-tuning on K400 and SSv2.}
LookWhen has more parameters but sparser computation, so it is also \emph{more memory-efficient} (please see Appendix \S \ref{app:speed_and_memory}).
In ``G$\times$T$\times$S'', G is the GFLOPs per view, T is the number of temporal views, S is the number of spatial views; their product is the total GFLOPs per video.
``\unk'' is unknown.
}
\label{fig:k400_ssv2_table}
\end{minipage}
\end{figure*}

\textbf{Setup: Comparing with existing models on K400 and SSv2.}
We compare LookWhen to existing ViT-B, Mamba-M, and Swin-B video models (since they are of similar size).
We include five adaptive-computation methods: LITE \cite{lite}, ToMe \cite{tome}, RLT \cite{rlt}, vid-TLDR \cite{vid_tldr}, and STTS \cite{stts}.
We include some larger models for reference but not direct comparison.
For both K400 and SSv2, we fine-tune our pre-trained LookWhen extractor for 30 epochs with our selector frozen.
We sweep 4 learning rates for each sparsity level \{70\%, 90\%\} to show performance at two operating points.

\textbf{Results: LookWhen is more compute-efficient than existing models (Tab. \ref{fig:k400_ssv2_table}).}
LookWhen achieves 84.6\% on K400 at 70\% sparsity, which is only 1.1\% lower than UMT-B800e \cite{umt} but LookWhen achieves it at 40\% fewer FLOPs.
The next best is VideoMambaPro \cite{lu2025snakes}, which scores 84.0\%; LookWhen is 0.6\% \emph{more} accurate at 78\% \emph{fewer} FLOPs.
On SSv2, LookWhen also achieves the best accuracy-compute trade-off, e.g. LookWhen scores 72.0\% at 108 GFLOPs per view, while VideoMAEv2 \cite{videomaev2} scores 71.2\% at 180 GFLOPs per view.

\begin{figure*}[t!]
\centering
\begin{subfigure}[b]{0.24\textwidth}
    \begin{tikzpicture}
\begin{axis}[
    width=1.25*\linewidth,
    height=\linewidth,
    title={K400, LP},
    title style={yshift=-8pt, font=\small},
    xlabel={Inference FLOPs (G)},
    ylabel={Accuracy (\%)},
    xlabel style={yshift=5pt, font=\scriptsize},
    ylabel style={yshift=-5pt, font=\scriptsize},
    tick label style={font=\scriptsize},
    xmin=0, xmax=270,
    ymin=71, ymax=83.3,
    grid=major,
    mark size=1.5pt,
    line width=1.2pt,
]

\addplot[color=lwpurple, mark=*, mark options={solid}]
coordinates {
    (25.2, 72.41)
    (40.05, 76.96)
    (55.7, 78.63)
    (108.3, 80.06)
    (127.68, 80.17)
    (147.92, 80.47)
    (191.03, 80.69)
};

\addplot[color=tldrteal, mark=square*, mark options={solid}]
coordinates {
    (44.3, 76.84)
    (81.1, 77.32)
    (137.9, 78.57)
    (169.0, 79.07)
    (191.2, 79.29)
};

\addplot[color=rltamber, mark=triangle*, mark options={solid}]
coordinates {
    (162, 78.5)
    (191, 78.66)
    (212, 78.78)
};

\node[
    star,
    star points=5,
    star point ratio=2,
    draw=black,
    fill=white,
    inner sep=1pt,
] at (axis cs:251.3, 81.85) {};

\end{axis}
\end{tikzpicture}
    \phantomcaption
    \label{fig:ft:k400}
\end{subfigure}
\hfill
\begin{subfigure}[b]{0.24\textwidth}
    \begin{tikzpicture}
\begin{axis}[
    width=1.25*\linewidth,
    height=\linewidth,
    title={SSv2, LP},
    title style={yshift=-8pt, font=\small},
    xlabel={Inference FLOPs (G)},
    ylabel={Accuracy (\%)},
    xlabel style={yshift=5pt, font=\scriptsize},
    ylabel style={yshift=-5pt, font=\scriptsize},
    tick label style={font=\scriptsize},
    xmin=0, xmax=270,
    ymin=34, ymax=52,
    grid=major,
    mark size=1.5pt,
    line width=1.2pt,
]

\addplot[color=lwpurple, mark=*, mark options={solid}]
coordinates {
    (25.2, 46.62)
    (40.05, 48.58)
    (55.7, 49.27)
    (108.3, 49.25)
    (127.68, 49.71)
    (147.92, 49.57)
    (191.03, 49.56)
};

\addplot[color=tldrteal, mark=square*, mark options={solid}]
coordinates {
    (44.3, 38.64)
    (91.3, 39.26)
    (137.9, 40.74)
    (169.0, 40.97)
    (191.2, 41.11)
};

\addplot[color=rltamber, mark=triangle*, mark options={solid}]
coordinates {
    (117, 35.36)
    (155, 35.69)
    (189, 35.37)
};

\node[
    star,
    star points=5,
    star point ratio=2,
    draw=black,
    fill=white,
    inner sep=1pt,
] at (axis cs:251.3, 40.95) {};

\end{axis}
\end{tikzpicture}
    \phantomcaption
    \label{fig:lp:k400}
\end{subfigure}
\hfill
\begin{subfigure}[b]{0.24\textwidth}
    \begin{tikzpicture}
\begin{axis}[
    width=1.25*\linewidth,
    height=\linewidth,
    title={K400, FT},
    title style={yshift=-8pt, font=\small},
    xlabel={Inference FLOPs (G)},
    ylabel={Accuracy (\%)},
    xlabel style={yshift=5pt, font=\scriptsize},
    ylabel style={yshift=-5pt, font=\scriptsize},
    tick label style={font=\scriptsize},
    xmin=0, xmax=270,
    ymin=74, ymax=87,
    grid=major,
    mark size=1.5pt,
    line width=1.2pt,
]

\addplot[color=lwpurple, mark=*, mark options={solid}]
coordinates {
    (25.2, 75.02)
    (40.05, 79.56)
    (55.7, 81.32)
    (108.3, 83.31)
    (127.68, 84.39)
    (147.92, 84.76)
    (191.03, 84.46)
};

\addplot[color=tldrteal, mark=square*, mark options={solid}]
coordinates {
    (44.3, 80.44)
    (91.3, 82.73)
    (137.9, 84.6)
    (164.0, 85.08)
    (185.9, 85.21)
};

\addplot[color=rltamber, mark=triangle*, mark options={solid}]
coordinates {
    (162, 84.35)
    (191, 84.37)
    (212, 84.45)
};

\node[
    star,
    star points=5,
    star point ratio=2,
    draw=black,
    fill=white,
    inner sep=1pt,
] at (axis cs:251.3, 85.29) {};

\end{axis}
\end{tikzpicture}
    \phantomcaption
    \label{fig:ft:ssv2}
\end{subfigure}
\hfill
\begin{subfigure}[b]{0.24\textwidth}
    \begin{tikzpicture}
\begin{axis}[
    width=1.25*\linewidth,
    height=\linewidth,
    title={SSv2, FT},
    title style={yshift=-8pt, font=\small},
    xlabel={Inference FLOPs (G)},
    ylabel={Accuracy (\%)},
    xlabel style={yshift=5pt, font=\scriptsize},
    ylabel style={yshift=-5pt, font=\scriptsize},
    tick label style={font=\scriptsize},
    xmin=0, xmax=270,
    ymin=62, ymax=72,
    grid=major,
    mark size=1.5pt,
    line width=1.2pt,
]

\addplot[color=lwpurple, mark=*, mark options={solid}]
coordinates {
    (25.2, 63.05)
    (40.05, 66.66)
    (55.7, 67.69)
    (108.3, 69.05)
    (127.68, 69.94)
    (147.92, 70.01)
    (191.03, 70.17)
};

\addplot[color=tldrteal, mark=square*, mark options={solid}]
coordinates {
    (44.3, 65.56)
    (100.2, 68.51)
    (137.9, 70.05)
    (164.0, 70.58)
    (191.2, 70.59)
};

\addplot[color=rltamber, mark=triangle*, mark options={solid}]
coordinates {
    (117, 69.7)
    (155, 70.23)
    (189, 70.3)
};

\node[
    star,
    star points=5,
    star point ratio=2,
    draw=black,
    fill=white,
    inner sep=1pt,
] at (axis cs:251.3, 70.86) {};

\end{axis}
\end{tikzpicture}
    \phantomcaption
    \label{fig:lp:ssv2}
\end{subfigure}
\\[-3ex]
\begin{subfigure}[b]{0.24\textwidth}
    \begin{tikzpicture}
\begin{axis}[
    width=1.25*\linewidth,
    height=\linewidth,
    title={Diving48, LP},
    title style={yshift=-8pt, font=\small},
    xlabel={Inference FLOPs (G)},
    ylabel={Accuracy (\%)},
    xlabel style={yshift=5pt, font=\scriptsize},
    ylabel style={yshift=-5pt, font=\scriptsize},
    tick label style={font=\scriptsize},
    xmin=0, xmax=270,
    ymin=19, ymax=43,
    grid=major,
    mark size=1.5pt,
    line width=1.2pt,
]

\addplot[color=lwpurple, mark=*, mark options={solid}]
coordinates {
    (25.2, 27.77)
    (40.05, 33.3)
    (55.7, 35.23)
    (108.3, 36.8)
    (127.68, 39.49)
    (147.92, 39.8)
    (191.03, 39.85)
};

\addplot[color=tldrteal, mark=square*, mark options={solid}]
coordinates {
    (44.3, 22.54)
    (100.2, 24.77)
    (128.3, 26.55)
    (161.2, 26.6)
    (184.4, 27.01)
};

\addplot[color=rltamber, mark=triangle*, mark options={solid}]
coordinates {
    (195, 26.4)
    (219, 26.95)
    (232, 26.19)
};

\node[
    star,
    star points=5,
    star point ratio=2,
    draw=black,
    fill=white,
    inner sep=1pt,
] at (axis cs:251.3, 27.41) {};

\end{axis}
\end{tikzpicture}
    \phantomcaption
    \label{fig:ft:diving}
\end{subfigure}
\hfill
\begin{subfigure}[b]{0.24\textwidth}
    \begin{tikzpicture}
\begin{axis}[
    width=1.25*\linewidth,
    height=\linewidth,
    title={EK-100, LP},
    title style={yshift=-8pt, font=\small},
    xlabel={Inference FLOPs (G)},
    ylabel={Accuracy (\%)},
    xlabel style={yshift=5pt, font=\scriptsize},
    ylabel style={yshift=-5pt, font=\scriptsize},
    tick label style={font=\scriptsize},
    xmin=0, xmax=270,
    ymin=17, ymax=31,
    grid=major,
    mark size=1.5pt,
    line width=1.2pt,
]

\addplot[color=lwpurple, mark=*, mark options={solid}]
coordinates {
    (25.2, 20.78)
    (40.05, 24.92)
    (55.7, 27.0)
    (108.3, 28.4)
    (127.68, 29.2)
    (147.92, 29.35)
    (191.03, 29.79)
};

\addplot[color=tldrteal, mark=square*, mark options={solid}]
coordinates {
    (44.3, 21.42)
    (81.1, 21.76)
    (128.3, 23.21)
    (169.0, 23.97)
    (184.4, 24.12)
};

\addplot[color=rltamber, mark=triangle*, mark options={solid}]
coordinates {
    (158, 18.81)
    (200, 19.14)
    (228, 19.02)
};

\node[
    star,
    star points=5,
    star point ratio=2,
    draw=black,
    fill=white,
    inner sep=1pt,
] at (axis cs:251.3, 24.12) {};

\end{axis}
\end{tikzpicture}
    \phantomcaption
    \label{fig:lp:diving}
\end{subfigure}
\hfill
\begin{subfigure}[b]{0.24\textwidth}
    \begin{tikzpicture}
\begin{axis}[
    width=1.25*\linewidth,
    height=\linewidth,
    title={Diving48, FT},
    title style={yshift=-8pt, font=\small},
    xlabel={Inference FLOPs (G)},
    ylabel={Accuracy (\%)},
    xlabel style={yshift=5pt, font=\scriptsize},
    ylabel style={yshift=-5pt, font=\scriptsize},
    tick label style={font=\scriptsize},
    xmin=0, xmax=270,
    ymin=61, ymax=94,
    grid=major,
    mark size=1.5pt,
    line width=1.2pt,
]

\addplot[color=lwpurple, mark=*, mark options={solid}]
coordinates {
    (25.2, 81.37)
    (40.05, 86.04)
    (55.7, 88.48)
    (108.3, 90.0)
    (127.68, 89.34)
    (147.92, 89.49)
    (191.03, 89.7)
};

\addplot[color=tldrteal, mark=square*, mark options={solid}]
coordinates {
    (44.3, 64.52)
    (91.3, 73.6)
    (128.6, 76.65)
    (164.0, 76.85)
    (185.9, 77.16)
};

\addplot[color=rltamber, mark=triangle*, mark options={solid}]
coordinates {
    (195, 85.99)
    (219, 85.79)
    (232, 85.89)
};

\node[
    star,
    star points=5,
    star point ratio=2,
    draw=black,
    fill=white,
    inner sep=1pt,
] at (axis cs:251.3, 85.79) {};

\end{axis}
\end{tikzpicture}
    \phantomcaption
    \label{fig:ft:ek}
\end{subfigure}
\hfill
\begin{subfigure}[b]{0.24\textwidth}
    \begin{tikzpicture}
\begin{axis}[
    width=1.25*\linewidth,
    height=\linewidth,
    title={EK-100, FT},
    title style={yshift=-8pt, font=\small},
    xlabel={Inference FLOPs (G)},
    ylabel={Accuracy (\%)},
    xlabel style={yshift=5pt, font=\scriptsize},
    ylabel style={yshift=-5pt, font=\scriptsize},
    tick label style={font=\scriptsize},
    xmin=0, xmax=270,
    ymin=37, ymax=55,
    grid=major,
    mark size=1.5pt,
    line width=1.2pt,
]

\addplot[color=lwpurple, mark=*, mark options={solid}]
coordinates {
    (25.2, 39.24)
    (40.05, 44.63)
    (55.7, 46.84)
    (108.3, 49.95)
    (127.68, 51.8)
    (147.92, 52.1)
    (191.03, 52.83)
};

\addplot[color=tldrteal, mark=square*, mark options={solid}]
coordinates {
    (44.3, 45.64)
    (100.2, 49.62)
    (137.9, 51.74)
    (161.2, 52.27)
    (184.4, 52.7)
};

\addplot[color=rltamber, mark=triangle*, mark options={solid}]
coordinates {
    (158, 50.93)
    (200, 51.12)
    (228, 51.25)
};

\node[
    star,
    star points=5,
    star point ratio=2,
    draw=black,
    fill=white,
    inner sep=1pt,
] at (axis cs:251.3, 52.99) {};

\end{axis}
\end{tikzpicture}
    \phantomcaption
    \label{fig:lp:ek}
\end{subfigure}
\\[-3ex]
\begin{subfigure}[b]{0.24\textwidth}
    \begin{tikzpicture}
\begin{axis}[
    width=1.25*\linewidth,
    height=\linewidth,
    title={Jester, LP},
    title style={yshift=-8pt, font=\small},
    xlabel={Inference FLOPs (G)},
    ylabel={Accuracy (\%)},
    xlabel style={yshift=5pt, font=\scriptsize},
    ylabel style={yshift=-5pt, font=\scriptsize},
    tick label style={font=\scriptsize},
    xmin=0, xmax=270,
    ymin=61, ymax=90,
    grid=major,
    mark size=1.5pt,
    line width=1.2pt,
]

\addplot[color=lwpurple, mark=*, mark options={solid}]
coordinates {
    (25.2, 81.21)
    (40.05, 84.04)
    (55.7, 84.64)
    (108.3, 85.21)
    (127.68, 85.32)
    (147.92, 85.62)
    (191.03, 85.39)
};

\addplot[color=tldrteal, mark=square*, mark options={solid}]
coordinates {
    (44.3, 65.29)
    (100.2, 71.87)
    (137.9, 74.52)
    (169.0, 75.17)
    (191.2, 75.42)
};

\addplot[color=rltamber, mark=triangle*, mark options={solid}]
coordinates {
    (68, 66.64)
    (95, 66.17)
    (134, 64.75)
};

\node[
    star,
    star points=5,
    star point ratio=2,
    draw=black,
    fill=white,
    inner sep=1pt,
] at (axis cs:251.3, 74.61) {};

\end{axis}
\end{tikzpicture}
    \phantomcaption
    \label{fig:ft:jester}
\end{subfigure}
\hfill
\begin{subfigure}[b]{0.24\textwidth}
    \begin{tikzpicture}
\begin{axis}[
    width=1.25*\linewidth,
    height=\linewidth,
    title={Charades, LP},
    title style={yshift=-8pt, font=\small},
    xlabel={Inference FLOPs (G)},
    ylabel={Accuracy (\%)},
    xlabel style={yshift=5pt, font=\scriptsize},
    ylabel style={yshift=-5pt, font=\scriptsize},
    tick label style={font=\scriptsize},
    xmin=0, xmax=270,
    ymin=24, ymax=38,
    grid=major,
    mark size=1.5pt,
    line width=1.2pt,
]

\addplot[color=lwpurple, mark=*, mark options={solid}]
coordinates {
    (25.2, 27.63)
    (40.05, 31.33)
    (55.7, 33.07)
    (108.3, 34.36)
    (127.68, 35.53)
    (147.92, 35.66)
    (191.03, 35.82)
};

\addplot[color=tldrteal, mark=square*, mark options={solid}]
coordinates {
    (44.3, 27.41)
    (100.2, 30.16)
    (137.9, 30.82)
    (169.0, 30.8)
    (191.2, 30.69)
};

\addplot[color=rltamber, mark=triangle*, mark options={solid}]
coordinates {
    (138, 25.51)
    (164, 25.53)
    (190, 25.4)
};

\node[
    star,
    star points=5,
    star point ratio=2,
    draw=black,
    fill=white,
    inner sep=1pt,
] at (axis cs:251.3, 30.35) {};

\end{axis}
\end{tikzpicture}
    \phantomcaption
    \label{fig:lp:jester}
\end{subfigure}
\hfill
\begin{subfigure}[b]{0.24\textwidth}
    \begin{tikzpicture}
\begin{axis}[
    width=1.25*\linewidth,
    height=\linewidth,
    title={Jester, FT},
    title style={yshift=-8pt, font=\small},
    xlabel={Inference FLOPs (G)},
    ylabel={Accuracy (\%)},
    xlabel style={yshift=5pt, font=\scriptsize},
    ylabel style={yshift=-5pt, font=\scriptsize},
    tick label style={font=\scriptsize},
    xmin=0, xmax=270,
    ymin=92, ymax=98,
    grid=major,
    mark size=1.5pt,
    line width=1.2pt,
]

\addplot[color=lwpurple, mark=*, mark options={solid}]
coordinates {
    (25.2, 95.91)
    (40.05, 96.67)
    (55.7, 96.83)
    (108.3, 96.98)
    (127.68, 97.04)
    (147.92, 97.04)
    (191.03, 97.03)
};

\addplot[color=tldrteal, mark=square*, mark options={solid}]
coordinates {
    (44.3, 92.59)
    (100.2, 95.54)
    (137.9, 96.23)
    (164.0, 96.38)
    (191.2, 96.42)
};

\addplot[color=rltamber, mark=triangle*, mark options={solid}]
coordinates {
    (68, 96.23)
    (95, 96.42)
    (134, 96.44)
};

\node[
    star,
    star points=5,
    star point ratio=2,
    draw=black,
    fill=white,
    inner sep=1pt,
] at (axis cs:251.3, 96.72) {};

\end{axis}
\end{tikzpicture}
    \phantomcaption
    \label{fig:ft:charades}
\end{subfigure}
\hfill
\begin{subfigure}[b]{0.24\textwidth}
    \begin{tikzpicture}
\begin{axis}[
    width=1.25*\linewidth,
    height=\linewidth,
    title={Charades, FT},
    title style={yshift=-8pt, font=\small},
    xlabel={Inference FLOPs (G)},
    ylabel={Accuracy (\%)},
    xlabel style={yshift=5pt, font=\scriptsize},
    ylabel style={yshift=-5pt, font=\scriptsize},
    tick label style={font=\scriptsize},
    xmin=0, xmax=270,
    ymin=32, ymax=50,
    grid=major,
    mark size=1.5pt,
    line width=1.2pt,
]

\addplot[color=lwpurple, mark=*, mark options={solid}]
coordinates {
    (25.2, 34.42)
    (40.05, 39.77)
    (55.7, 42.2)
    (108.3, 44.8)
    (127.68, 46.47)
    (147.92, 46.77)
    (191.03, 47.17)
};

\addplot[color=tldrteal, mark=square*, mark options={solid}]
coordinates {
    (44.3, 37.7)
    (100.2, 42.27)
    (137.9, 44.26)
    (169.0, 44.95)
    (191.2, 45.09)
};

\addplot[color=rltamber, mark=triangle*, mark options={solid}]
coordinates {
    (138, 44.2)
    (164, 44.45)
    (190, 44.62)
};

\node[
    star,
    star points=5,
    star point ratio=2,
    draw=black,
    fill=white,
    inner sep=1pt,
] at (axis cs:251.3, 45.11) {};

\end{axis}
\end{tikzpicture}
    \phantomcaption
    \label{fig:lp:charades}
\end{subfigure}
\\[-3ex]
\caption{\textbf{Linear probing (LP) and fine-tuning (FT) accuracy vs.\ FLOPs} across six datasets.
Our LookWhen (\textcolor{lwpurple}{\raisebox{-0.3pt}{\Large$\bullet$}}) mostly outperforms the baselines in controlled settings.
Gains are largest for LP, sometimes surpassing the \emph{dense} InternVideo2 ($\star$).
We make these upgraded baselines by applying the sparsification methods vid-TLDR (\textcolor{tldrteal}{$\blacksquare$}) \cite{vid_tldr} or RLT (\textcolor{rltamber}{$\blacktriangle$}) \cite{rlt} to the SOTA ViT-B InternVideo2 \cite{internvideo2}.
}
\label{fig:lp_ft_efficiency}
\end{figure*}
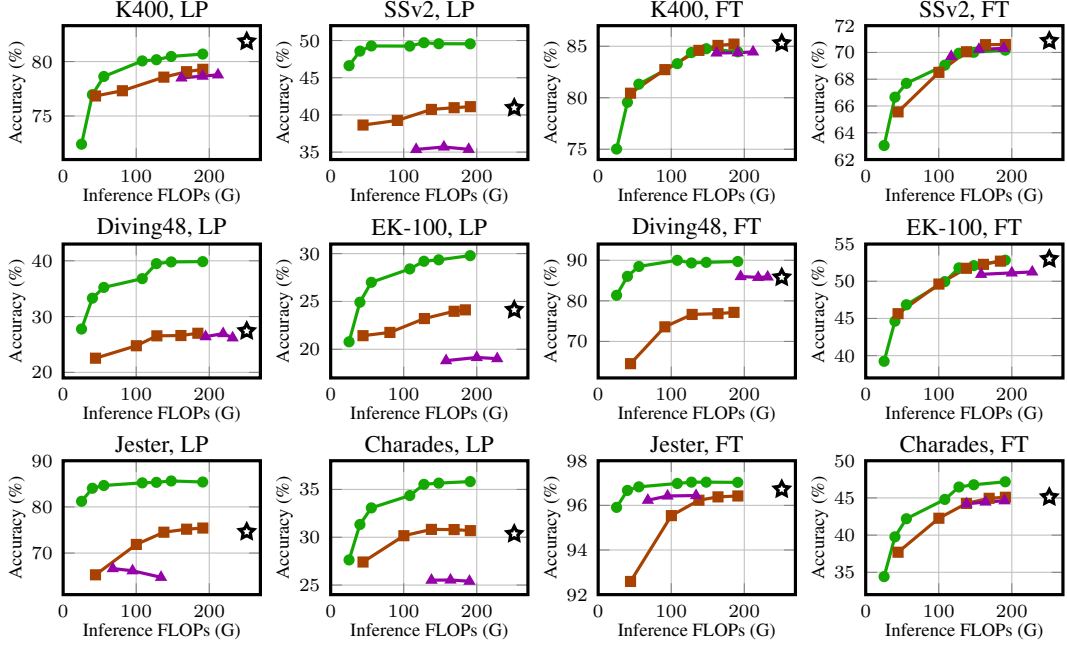

\textbf{Setup: Controlled and upgraded baselines.}
InternVideo2-B (IV2 for short) warrants close comparison to LookWhen as our video teacher.
However, there are no published results at this scale (results only exist for its much larger variants), so we run it ourselves.
We also apply two adaptive computation methods, vid-TLDR \cite{vid_tldr} and RLT \cite{rlt}, to the IV2 backbone as efficient editions of our teacher.
For IV2, IV2+RLT, and our LookWhen, we fine-tune for 10 epochs on K400 \cite{k400}, SSv2 \cite{ssv2}, Jester \cite{jester}, and Epic-Kitchens-100 \cite{ek-100}; and for 50 epochs for the smaller Diving48 \cite{diving48} and Charades \cite{charades} datasets.
We equally tune the learning rate for all methods to be fair.
Since vid-TLDR sparsifies an already fine-tuned model, we apply it to our fine-tuned IV2 models.
We care about the accuracy-cost \emph{trade-off}, so we measure performance at multiple operating points by varying sparsity.
During fine-tuning, we vary RLT's threshold $\tau \in \{0.05, 0.1, 0.2\}$ and LookWhen's sparsity between 50\% and 95\%.
The vid-TLDR paper tries many token-merging schedules for each dataset, so we do the same by trying 4 different schedules per sparsity level and choose the best per level per dataset.

\textbf{Results: LookWhen is more efficient than upgraded baselines (Figure~\ref{fig:lp_ft_efficiency}).}
LookWhen Pareto-dominates on 9 of 12 evaluations and roughly matches on the other 3.
LookWhen shines in feature extraction, e.g. \textgreater10\% more accurate on Diving48 LP.
LookWhen sometimes even outperforms its dense video teacher, IV2. We attribute it to our DINOv3 video-token pre-training (see \S\ref{sec:ablations}).


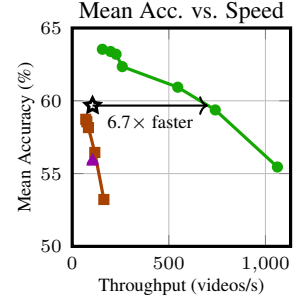
\begin{wrapfigure}[16]{r}{0.32\textwidth}
    \vspace{-\intextsep}
    \begin{tikzpicture}
\begin{axis}[
    width=\linewidth,
    height=\linewidth,
    title={Mean Acc. vs. Speed},
    title style={yshift=-8pt, font=\small},
    xlabel={Throughput (videos/s)},
    ylabel={Mean Accuracy (\%)},
    xlabel style={yshift=5pt, font=\scriptsize},
    ylabel style={yshift=-5pt, font=\scriptsize},
    tick label style={font=\scriptsize},
    xmin=0, xmax=1130,
    ymin=50, ymax=65,
    grid=major,
    mark size=1.5pt,
    line width=1.2pt,
]

\addplot[color=lwpurple, mark=*, mark options={solid}]
coordinates {
    (1061.94, 55.4525)
    (740.2, 59.37166667)
    (546.15, 60.93333333)
    (259.1, 62.3475)
    (226.89, 63.2)
    (198.57, 63.38666667)
    (156.11, 63.53833333)
};

\addplot[color=tldrteal, mark=square*, mark options={solid}]
coordinates {
    (163.82, 53.22)
    (116.89, 56.45)
    (83.97, 58.16)
    (75.57, 58.56)
    (70.19, 58.73)
};

\addplot[color=rltamber, mark=triangle*, mark options={solid}]
coordinates {
    (105.25, 55.7975)
    (105.25, 55.92666667)
    (105.25, 55.8225)
};

\node[
    star,
    star points=5,
    star point ratio=2,
    draw=black,
    fill=white,
    inner sep=1pt,
] at (axis cs:105.25, 59.67083333) {};

\draw[->, thick, black]
    (axis cs:105.25, 59.67083333) -- (axis cs:703, 59.67083333)
    node[midway, below, font=\scriptsize] {6.7$\times$ faster};

\end{axis}
\end{tikzpicture}
    \caption{LookWhen (\textcolor{lwpurple}{\raisebox{-0.3pt}{\Large$\bullet$}}) dominates baselines in mean accuracy (over 6 datasets and 2 settings) versus measured throughput.
    Markers: IV2 ($\star$), IV2+vid-TLDR (\textcolor{tldrteal}{$\blacksquare$}), and IV2+RLT (\textcolor{rltamber}{$\blacktriangle$}).}
    \label{fig:efficiency_in_practice}
\end{wrapfigure}

\textbf{\emph{Realized} efficiency: LookWhen's efficiency gains increase when measured in practice.}
We measure throughput on an NVIDIA L40S GPU to check if theoretical gains (accuracy-FLOPs) translates to practical gains (accuracy-throughput).
Our efficiency gains substantially increase; e.g. at equal accuracy, LookWhen is 6.7$\times$ faster than IV2 (Fig. \ref{fig:efficiency_in_practice}), where accuracy is the mean across the 6 datasets and 2 settings (LP and FT).
IV2+vid-TLDR is less efficient in practice because of Flash Attention \cite{dao2023flashattention2} incompatibility for token-merging layers (the first 2-4 layers).
IV2+RLT is less efficient in practice because it requires a batch size of 1 or masking (where all tokens are processed and the contributions of some tokens are ignored).
Our LookWhen uses standard deep learning operations so is efficient in practice using standard libraries, e.g. PyTorch \cite{paszke2019pytorch}.
Because of its selector, LookWhen has more parameters than IV2, yet LookWhen is \emph{more memory-efficient} because it has fewer activations.
Please see accuracy-speed plots for all 12 evaluations and memory-use statistics in Appendix \S \ref{app:speed_and_memory}.

\section{Ablations and Analysis: Looking Inside}\label{sec:ablations}

We analyze the key design choices behind LookWhen. We first describe the ablation dataset and evaluation setup, then study the selection and extraction targets.
Full experimental details, including sampling, augmentations, and hyperparameters, are provided in Appendix~\ref{appendix:implementation}.


\textbf{Datasets.}
We re-use the six tasks to assess general performance.
For feasibility, we use 20K training subsets for all except Diving48 and Charades, which have less than 20K, and use all validation points.

\textbf{Evaluating.}
To assess both feature quality and fine-tuning ability, we report results for both linear probing (LP) and fine-tuning (FT). 
For LP, we train a linear classifier on frozen features. For FT, we fine-tune the extractor and head, initializing the head from the LP checkpoint. 
LP reflects performance in low-compute settings, while FT captures performance when more accuracy is desired.
To make it simple, we evaluate at 90\% sparsity for these main-paper and Appendix (\S\ref{appendix:more_ablations}) ablations.

\begin{table*}[t]
    \centering
    \caption{
\textbf{Ablating \emph{when} and \emph{where} to select.}
We train with different selection targets to study its effect on downstream tasks.
LP is linear probing the frozen extractor, FT is fine-tuning the extractor.
All runs use InternVideo2 video-token target, frame and patch loss (no norm), and K400+SSv2 data. 
    }
    \label{tab:super_table}
    \begin{minipage}{\linewidth}
        \subcaption{
\textbf{Attention vs. token uniqueness.}
Training to select unique tokens (top1-dist) beats highly-attended tokens, if we have suitable teacher features (e.g. DINOv3).
The \emph{change} in DINOv3's attention between successive frames ($\Delta$) beats InternVideo2's space-time attention (which contains artifacts) and DINOv3's space-only attention.
        }
        \label{tab:selector_target}
\centering
\scriptsize
\setlength{\tabcolsep}{6.6pt}
\label{tab:attn_vs_rep}
\begin{tabular}{llcccccccccccc}
&
& \multicolumn{2}{c}{K400-20K}
& \multicolumn{2}{c}{SSv2-20K}
& \multicolumn{2}{c}{Diving48}
& \multicolumn{2}{c}{EK100-20K}
& \multicolumn{2}{c}{Jester-20K}
& \multicolumn{2}{c}{Charades}
\\
\cmidrule(lr){3-4} \cmidrule(lr){5-6} \cmidrule(lr){7-8} \cmidrule(lr){9-10}  \cmidrule(lr){11-12}  \cmidrule(lr){13-14}
Teacher
& Method
& LP & FT
& LP & FT
& LP & FT
& LP & FT
& LP & FT
& LP & FT
\\
\midrule
\rowcolor{when!20} DINOv3 & top1-dist &
\underline{72.9} & \underline{73.6} & \underline{31.5} & \underline{45.2} & \textbf{29.7} & \textbf{83.2} & \textbf{11.6} & \textbf{22.2} & \textbf{63.4} & \textbf{94.7} & \textbf{29.3} & \textbf{36.7}\\
DINOv3 & attn &
72.1 & 72.8 & 30.1 & 41.1 & \underline{28.0} & \underline{77.7} & 9.1 & 17.6 & 53.8 & 90.1 & 26.6 & 33.2\\
DINOv3 & $\Delta$attn &
\textbf{73.5} & \textbf{74.2} & \textbf{31.7} & \textbf{45.6} & 27.6 & 76.0 & \underline{11.0} & \underline{21.3} & \textbf{63.4} & \underline{94.4} & \underline{28.6} & \underline{35.7}\\
InternVideo2 & top1-dist &
64.7 & 64.7 & 27.0 & 31.2 & 15.8 & 44.6 & 7.3 & 13.4 & 56.8 & 89.2 & 21.9 & 26.0\\
InternVideo2 & attn &
68.6 & 68.9 & 28.6 & 35.4 & 26.2 & 72.9 & 9.3 & 15.2 & \underline{58.2} & 89.9 & 24.4 & 30.4\\
none & random &
72.5 & 72.7 & 30.4 & 42.8 & 18.3 & 57.2 & 10.2 & 20.3 & 57.2 & 91.8 & 25.9 & 30.8\\

\end{tabular}
    \end{minipage}
    \vspace{1em}
    \begin{minipage}{\linewidth}
        \subcaption{
\textbf{Computing token uniqueness.}
We vary K in our ``topK-distance'' method, which is each patch's mean distance to its K-nearest neighbors in DINOv3's feature space.
K=1 performs the best on average.
We also try ``K-center'' on both features and pixels \cite{k_center}, which runs farthest-point sampling \cite{furthest_point} for maximum diversity.
        }
        \label{tab:selector_target_k}
\centering
\scriptsize
\setlength{\tabcolsep}{7.8pt}
\label{tab:finding_unique}
\begin{tabular}{lcccccccccccc}
& \multicolumn{2}{c}{K400-20K}
& \multicolumn{2}{c}{SSv2-20K}
& \multicolumn{2}{c}{Diving48}
& \multicolumn{2}{c}{EK100-20K}
& \multicolumn{2}{c}{Jester-20K}
& \multicolumn{2}{c}{Charades}
\\
\cmidrule(lr){2-3} \cmidrule(lr){4-5} \cmidrule(lr){6-7} \cmidrule(lr){8-9}  \cmidrule(lr){10-11}  \cmidrule(lr){12-13}
Method
& LP & FT
& LP & FT
& LP & FT
& LP & FT
& LP & FT
& LP & FT
\\
\midrule
\rowcolor{when!20} top1-distance &
\textbf{72.9} & \textbf{73.6} & 31.5 & \textbf{45.2} & 29.7 & 83.2 & \textbf{11.6} & \textbf{22.2} & \underline{63.4} & \underline{94.7} & \textbf{29.3} & \textbf{36.7}\\
top10-distance &
\underline{72.7} & \underline{73.2} & \underline{31.6} & 44.7 & \underline{30.5} & \textbf{84.0} & \underline{11.3} & \underline{21.6} & 63.3 & \textbf{94.8} & \underline{28.9} & \underline{36.4}\\
top100-distance &
71.4 & 72.3 & 31.4 & 43.5 & \textbf{31.0} & \underline{83.5} & 11.0 & 20.5 & 61.8 & 94.4 & 27.8 & 35.4\\
feature K-center &
\textbf{72.9} & \textbf{73.6} & \textbf{31.8} & \underline{44.9} & 27.1 & 79.5 & 10.7 & 20.8 & 63.1 & \textbf{94.8} & 28.2 & 35.4\\
pixel K-center &
72.1 & 72.8 & 31.4 & 42.9 & 25.9 & 72.7 & 9.8 & 16.7 & \textbf{66.0} & 93.8 & 28.6 & 35.3\\
\end{tabular}
    \end{minipage}
\end{table*}

\subsection{\emph{When} and \emph{where} to \textcolor{lowcolor}{\textbf{select}}}

\textbf{Attention vs. token uniqueness (Tab. \ref{tab:selector_target}).}
We first explore selection training paradigms: attention-based or representation-based.
Directly comparing our method (top1-dist using DINOv3) to a naive extension of Look\emph{Where} \cite{lookwhere} to video (attention using IV2), our method wins easily: +3.9\% LP and +6.7\% FT.
Training to select IV2's attention is no better than random selection, showing that IV2's noisy space-time attention cannot effectively guide selection.
Plugging our top1-dist into IV2's feature map also fails---it is much worse than random selection.
This aligns with our intuition that positionally-grounded representations or attention is needed for effective selection.
We also tried to exploit time redundancy via ``$\Delta$attn'' (row \#3), which measures how rapidly each patch's \emph{DINOv3} attention changes across neighboring frames.
This method works well-enough but underperforms our top1-dist overall by 0.4\% LP and 1.4\% FT.
In particular, $\Delta$attn is least effective on Diving48, which follows divers while panning the camera---our top1-distance is robust to this challenging motion. 

\textbf{Measuring token uniqueness (Tab. \ref{tab:selector_target_k}).}
We explore different variations of our top1-dist method, both in DINOv3's feature space (rows \#1-4) and the pixel space (row \#5).
Computing the average distance to more neighbors still works well (rows \#2,3), but 1 neighbor is best (row \#1).
We also try further increasing the diversity of token-selection targets via farthest-point sampling \cite{furthest_point} in DINOv3's feature space (row \#4) and the pixel space (row \#5); the latter is used in prior work \cite{k_center}.




\subsection{\emph{What} to \textcolor{highcolor}{\textbf{extract}}}

\textbf{Video-token targets (Tab.~\ref{tab:extractor_video_target}).}
We first explore video-token supervision with frame-level and patch-level losses disabled.
Video-token distillation with IV2 (row \#3) significantly outperforms distillation with DINOv3 (row \#4) by +12.8\% in LP and +2.5\% in FT on average.
Yet DINOv3 alone beats IV2 alone in some cases (+8.8\% on Jester LP).
Including \emph{both} targets (row \#1) boosts gains over IV2 alone (+3\% in LP and 0.7\% in FT on average), confirming that the two forms of supervision are complementary.
\emph{Removing} within-video target-feature normalization of DINOv3 (row \#2) drops performance (-2.1\% LP and -0.4\% FT on average) and sometimes by large margins (-6.4\% on SSv2 LP and -5.1\% on Jester LP), showing that learning \emph{what changes} is an effective strategy.


\textbf{Frame- and patch-token targets (Tab.~\ref{tab:extractor_patch_target}).}
We explore frame-level and patch-level supervision with both video-token losses enabled.
Supervising both frame and patch tokens (row \#1) outperforms supervising neither (row \#4) by +2.1\% LP and +1.3\% FT on average; these gains are sometimes massive, e.g. +7.7\% on Diving48 LP and +10\% on Jester LP.
Supervising either frame (row \#2) or patch tokens (row \#3) narrows the gap.
We note that supervising all tokens is not uniformly optimal, e.g. K400 is best without these denser supervision signals.
See Appendix \S\ref{appendix:more_ablations} for more ablations.

\begin{table*}[t]
    \centering
    \caption{
\textbf{Ablating \emph{what} to extract.}
We train with different extractor targets to study its effect on downstream tasks.
LP is linear probing the frozen extractor, FT is fine-tuning the extractor.
All runs use top1-distance using DINOv3's feature space to train the selector map and K400+SSv2 datasets.
    }
    \label{tab:super_table}
    \begin{minipage}{\linewidth}
        \subcaption{
\textbf{Video-token target.}
Predicting \emph{both} InternVideo2's video token and a video token we make from DINOv3's frame-wise class tokens improves representations.
We time-normalize each dimension before concatenating DINOv3 tokens to learn \emph{what} changes.
Frame and patch losses are disabled to isolate video-token supervision.
        }
        \label{tab:extractor_video_target}
\centering
\scriptsize
\setlength{\tabcolsep}{5pt}

\begin{tabular}{ccccccccccccccccc}
& \multicolumn{4}{c}{DINOv3}
& \multicolumn{2}{c}{K400-20K}
& \multicolumn{2}{c}{SSv2-20K}
& \multicolumn{2}{c}{Diving48}
& \multicolumn{2}{c}{EK100-20K}
& \multicolumn{2}{c}{Jester-20K}
& \multicolumn{2}{c}{Charades}
\\
\cmidrule(lr){2-5}
\cmidrule(lr){6-7} \cmidrule(lr){8-9} \cmidrule(lr){10-11}  \cmidrule(lr){12-13}  \cmidrule(lr){14-15}  \cmidrule(lr){16-17}
IntVid2 & Vid & Frame & Patch & Norm
& LP & FT
& LP & FT
& LP & FT
& LP & FT
& LP & FT
& LP & FT
\\
\midrule
\rowcolor{when!20} \ding{52} & \ding{52} & \ding{56} & \ding{56} & \ding{52} &
\textbf{73.3} & \textbf{74.0} & \textbf{37.9} & \underline{48.0} & \underline{29.9} & 81.9 & \textbf{12.8} & \textbf{23.3} & \underline{69.0} & 94.6 & \textbf{30.3} & \textbf{37.4}\\
\ding{52} & \ding{52} & \ding{56} & \ding{56} & \ding{56} &
\underline{72.1} & 73.3 & 31.5 & 45.1 & \textbf{32.1} & \underline{84.2} & \underline{11.3} & \underline{22.4} & 63.9 & \underline{94.9} & \underline{29.5} & \underline{36.9}\\
\ding{52} & \ding{56} & \ding{56} & \ding{56} & \ding{56} &
\textbf{73.3} & \underline{73.9} & \underline{32.1} & 45.6 & 27.9 & 83.1 & 11.0 & 22.0 & 61.6 & 94.4 & 29.0 & 36.3\\
\ding{56} & \ding{52} & \ding{56} & \ding{56} & \ding{52} &
17.9 & 61.1 & 26.7 & \textbf{48.8} & 26.7 & \textbf{85.7} & 2.6 & 18.7 & \textbf{70.4} & \textbf{95.2} & 14.0 & 30.5\\
\end{tabular}

    \end{minipage}
    \vspace{1em}
    \begin{minipage}{\linewidth}
        \subcaption{
\textbf{Frame and patch-token targets.}
We turn off frame and patch distillation losses to study their effects within the full IV2+DINOv3 configuration.
Using both losses helps on average; removing both hurts Jester and Diving48 the most. All rows include normalization over time for constructing DINOv3 Vid representation.
        }
        \label{tab:extractor_patch_target}
\centering
\scriptsize
\setlength{\tabcolsep}{5.85pt}
\begin{tabular}{cccccccccccccccc}
& \multicolumn{3}{c}{DINOv3}
& \multicolumn{2}{c}{K400-20K}
& \multicolumn{2}{c}{SSv2-20K}
& \multicolumn{2}{c}{Diving48}
& \multicolumn{2}{c}{EK100-20K}
& \multicolumn{2}{c}{Jester-20K}
& \multicolumn{2}{c}{Charades}
\\
\cmidrule(lr){2-4}
\cmidrule(lr){5-6} \cmidrule(lr){7-8} \cmidrule(lr){9-10}  \cmidrule(lr){11-12}  \cmidrule(lr){13-14}  \cmidrule(lr){15-16}
IntVid2 & Vid & Frame & Patch
& LP & FT
& LP & FT
& LP & FT
& LP & FT
& LP & FT
& LP & FT
\\
\midrule
\rowcolor{when!20} \ding{52} & \ding{52} & \ding{52} & \ding{52} &
72.5 & \underline{73.8} & 38.1 & \textbf{50.1} & \textbf{37.6} & \textbf{84.8} & \underline{13.5} & \underline{23.8} & \textbf{79.0} & \underline{95.2} & \textbf{31.3} & \textbf{39.3}\\
\ding{52} & \ding{52} & \ding{52} & \ding{56} &
72.5 & \underline{73.8} & \underline{38.6} & \textbf{50.1} & \underline{35.5} & 84.1 & \textbf{13.8} & \textbf{24.2} & 77.0 & \underline{95.2} & \underline{31.0} & \underline{39.1}\\
\ding{52} & \ding{52} & \ding{56} & \ding{52} &
\underline{72.7} & \underline{73.8} & \textbf{39.1} & \underline{49.6} & 31.9 & \underline{84.6} & \textbf{13.8} & \underline{23.8} & \underline{77.1} & \textbf{95.3} & 30.7 & 38.7\\
\ding{52} & \ding{52} & \ding{56} & \ding{56} &
\textbf{73.3} & \textbf{74.0} & 37.9 & 48.0 & 29.9 & 81.9 & 12.8 & 23.3 & 69.0 & 94.6 & 30.3 & 37.4\\
\end{tabular}

    \end{minipage}
\end{table*}

\section{Related work: Looking around}



\textbf{Token reduction} methods drop or merge tokens across layers to reduce computation.
For example, ToMe \citep{tome} merges tokens via soft bipartite matching.
Other methods have not been extended from images to video, so we do not include them as baselines \cite{atc, dtem, pitome}.
STA \citep{sta} progressively prunes video tokens by accumulating inter-frame similarity over time and reweighting via activation-based semantic scores.
vid-TLDR \citep{vid_tldr} defines a saliency score based on attention sharpness and merges accordingly, enabling earlier reduction; it beats ToMe, so we use it as a baseline in our controlled experiments (\S \ref{experiments}).
Merging methods typically reduce FLOPs more than runtime, since their algorithms may not be GPU-friendly or Flash-Attention-compatible; hence LookWhen's gains over IV2+vid-TLDR grow in reality.
These methods still process all tokens in the first layer and only reduce them gradually, so their peak memory use can be very high (see memory-use statistics in \S \ref{app:speed_and_memory}).
In contrast, LookWhen never processes all tokens, is GPU-friendly, and is FLOP, runtime, \emph{and} memory efficient.


\textbf{Token selection} methods need to learn to efficiently select informative parts of the input. 
RLT \cite{rlt} uses pixel similarity to remove static patches.
K-centered \cite{k_center} samples patches to maximize pixel-level diversity.
EVEREST \cite{hwang2024everest} selects high-motion tokens via adjacent-frame embedding differences.
In contrast, LookWhen selects tokens based on \emph{feature-space} uniqueness, enabling more aggressive sparsity.
AutoGaze \cite{video-gaze}, a concurrent work, uses a lightweight model to autoregressively select a minimal set of patches that can reconstruct each frame within a fixed error threshold, removing redundant regions.
STTS \cite{stts} trains a scorer network end-to-end to rank token importance using a differentiable top-K operator. 
LITE \cite{lite} trains a selector using gradients of class scores with respect to feature activations.
Although LITE shares the spirit of LookWhen, the methods differ in important ways: LookWhen trains a joint selector-extractor rather than only a selector; its selector exploits general video redundancy rather than task-specific redundancy; and its extractor predicts general representations from two teachers, making it directly usable as an effective video feature extractor for linear probing as well as an effective model for fine-tuning.

\textbf{Architectural changes.}
Other work modifies token interactions, e.g. attention \citep{timesformer, space-time-mixing, video-focalnets} or the overall architecture \citep{x3d, slowfast, videomamba, memvit}.
They reduce the cost of token processing, but still operate on the full set of tokens: they process all tokens, less.
LookWhen processes fewer tokens, more.

\textbf{Multimodal models.}
Recent work makes vision-language models more efficient \citep{nvila, PuMer, storm, longvu, chat-univi, testa, slowfast-llava, llava-mini, dycoke}.
They reduce tokens \emph{after} vision encoder processing, sparing language model computation, which dominates the cost.
They are not designed for unimodal video recognition, while LookWhen spares vision computation.

\section{Closing: Looking to the end}

\textbf{Limitations and future work.}
Due to computational constraints, we are restricted to training ViT-Base models and could not include larger ViTs. 
One potential limitation of LookWhen’s design is that it requires a teacher with positionally-grounded patch-level attention or representation to effectively supervise the selector.
However, concurrent foundation model development now includes ongoing efforts to achieve this property for images \cite{eupe, khosla2026t, cao2026tipsv2} and video \cite{v_jepa_21}, due to the utility and transferability of positionally-grounded outputs.
As a result more compatible teachers may be available and soon.
Future efforts can explore long video (Look\emph{Then}), multi-view (Look\emph{How}), and hyperspectral (Look\emph{Which}) processing, all of which introduce significant computational costs and their own forms of redundancy and opportunities for selection.

\textbf{Conclusion.}
We extend the selector-extractor framework from images to video by (1) adapting the network architecture, (2) pre-training to select tokens that have unique features, and (3) pre-training to extract features from a video teacher \emph{and} from an image teacher to learn both global video-level and fine-grained patch-level features.
Through ablations we show that these strategies explain why our LookWhen achieves better accuracy-compute trade-offs than existing models and other baselines on six video benchmarks. 
When released, we hope the LookWhen code and pre-trained models can make video recognition computationally feasible for you.

\clearpage

\bibliographystyle{unsrt}
\bibliography{references.bib}

\clearpage
\appendix

\section{Technical Appendices and Supplementary Material}

\subsection{Efficiency in Practice: Throughput and Memory Measurements}\label{app:speed_and_memory}

\begin{figure*}[h]
\centering
\begin{subfigure}[b]{0.24\textwidth}
    \begin{tikzpicture}
\begin{axis}[
    width=1.25*\linewidth,
    height=\linewidth,
    title={K400, LP},
    title style={yshift=-8pt, font=\small},
    xlabel={Throughput (videos/s)},
    ylabel={Accuracy (\%)},
    xlabel style={yshift=5pt, font=\scriptsize},
    ylabel style={yshift=-5pt, font=\scriptsize},
    tick label style={font=\scriptsize},
    xmin=0, xmax=1130,
    ymin=71, ymax=83.3,
    grid=major,
    mark size=1.5pt,
    line width=1.2pt,
]

\addplot[color=lwpurple, mark=*, mark options={solid}]
coordinates {
    (1061.94, 72.41)
    (740.2, 76.96)
    (546.15, 78.63)
    (259.1, 80.06)
    (226.89, 80.17)
    (198.57, 80.47)
    (156.11, 80.69)
};

\addplot[color=tldrteal, mark=square*, mark options={solid}]
coordinates {
    (163.82, 76.84)
    (155.15, 77.32)
    (79.61, 78.57)
    (68.4, 79.07)
    (62.36, 79.29)
};

\addplot[color=rltamber, mark=triangle*, mark options={solid}]
coordinates {
    (105.25, 78.5)
    (105.25, 78.66)
    (105.25, 78.78)
};

\node[
    star,
    star points=5,
    star point ratio=2,
    draw=black,
    fill=white,
    inner sep=1pt,
] at (axis cs:105.25, 81.85) {};

\end{axis}
\end{tikzpicture}
    \phantomcaption
    \label{fig:ft:k400}
\end{subfigure}
\hfill
\begin{subfigure}[b]{0.24\textwidth}
    \begin{tikzpicture}
\begin{axis}[
    width=1.25*\linewidth,
    height=\linewidth,
    title={SSv2, LP},
    title style={yshift=-8pt, font=\small},
    xlabel={Throughput (videos/s)},
    ylabel={Accuracy (\%)},
    xlabel style={yshift=5pt, font=\scriptsize},
    ylabel style={yshift=-5pt, font=\scriptsize},
    tick label style={font=\scriptsize},
    xmin=0, xmax=1130,
    ymin=34, ymax=52,
    grid=major,
    mark size=1.5pt,
    line width=1.2pt,
]

\addplot[color=lwpurple, mark=*, mark options={solid}]
coordinates {
    (1061.94, 46.62)
    (740.2, 48.58)
    (546.15, 49.27)
    (259.1, 49.25)
    (226.89, 49.71)
    (198.57, 49.57)
    (156.11, 49.56)
};

\addplot[color=tldrteal, mark=square*, mark options={solid}]
coordinates {
    (163.82, 38.64)
    (123.49, 39.26)
    (79.61, 40.74)
    (68.4, 40.97)
    (62.36, 41.11)
};

\addplot[color=rltamber, mark=triangle*, mark options={solid}]
coordinates {
    (105.25, 35.36)
    (105.25, 35.69)
    (105.25, 35.37)
};

\node[
    star,
    star points=5,
    star point ratio=2,
    draw=black,
    fill=white,
    inner sep=1pt,
] at (axis cs:105.25, 40.95) {};

\end{axis}
\end{tikzpicture}
    \phantomcaption
    \label{fig:lp:k400}
\end{subfigure}
\hfill
\begin{subfigure}[b]{0.24\textwidth}
    \begin{tikzpicture}
\begin{axis}[
    width=1.25*\linewidth,
    height=\linewidth,
    title={K400, FT},
    title style={yshift=-8pt, font=\small},
    xlabel={Throughput (videos/s)},
    ylabel={Accuracy (\%)},
    xlabel style={yshift=5pt, font=\scriptsize},
    ylabel style={yshift=-5pt, font=\scriptsize},
    tick label style={font=\scriptsize},
    xmin=0, xmax=1130,
    ymin=74, ymax=87,
    grid=major,
    mark size=1.5pt,
    line width=1.2pt,
]

\addplot[color=lwpurple, mark=*, mark options={solid}]
coordinates {
    (1061.94, 75.02)
    (740.2, 79.56)
    (546.15, 81.32)
    (259.1, 83.31)
    (226.89, 84.39)
    (198.57, 84.76)
    (156.11, 84.46)
};

\addplot[color=tldrteal, mark=square*, mark options={solid}]
coordinates {
    (163.82, 80.44)
    (123.49, 82.73)
    (79.61, 84.6)
    (82.39, 85.08)
    (73.83, 85.21)
};

\addplot[color=rltamber, mark=triangle*, mark options={solid}]
coordinates {
    (105.25, 84.35)
    (105.25, 84.37)
    (105.25, 84.45)
};

\node[
    star,
    star points=5,
    star point ratio=2,
    draw=black,
    fill=white,
    inner sep=1pt,
] at (axis cs:105.25, 85.29) {};

\end{axis}
\end{tikzpicture}
    \phantomcaption
    \label{fig:ft:ssv2}
\end{subfigure}
\hfill
\begin{subfigure}[b]{0.24\textwidth}
    \begin{tikzpicture}
\begin{axis}[
    width=1.25*\linewidth,
    height=\linewidth,
    title={SSv2, FT},
    title style={yshift=-8pt, font=\small},
    xlabel={Throughput (videos/s)},
    ylabel={Accuracy (\%)},
    xlabel style={yshift=5pt, font=\scriptsize},
    ylabel style={yshift=-5pt, font=\scriptsize},
    tick label style={font=\scriptsize},
    xmin=0, xmax=1130,
    ymin=62, ymax=72,
    grid=major,
    mark size=1.5pt,
    line width=1.2pt,
]

\addplot[color=lwpurple, mark=*, mark options={solid}]
coordinates {
    (1061.94, 63.05)
    (740.2, 66.66)
    (546.15, 67.69)
    (259.1, 69.05)
    (226.89, 69.94)
    (198.57, 70.01)
    (156.11, 70.17)
};

\addplot[color=tldrteal, mark=square*, mark options={solid}]
coordinates {
    (163.82, 65.56)
    (103.13, 68.51)
    (79.61, 70.05)
    (82.39, 70.58)
    (62.36, 70.59)
};

\addplot[color=rltamber, mark=triangle*, mark options={solid}]
coordinates {
    (105.25, 69.7)
    (105.25, 70.23)
    (105.25, 70.3)
};

\node[
    star,
    star points=5,
    star point ratio=2,
    draw=black,
    fill=white,
    inner sep=1pt,
] at (axis cs:105.25, 70.86) {};

\end{axis}
\end{tikzpicture}
    \phantomcaption
    \label{fig:lp:ssv2}
\end{subfigure}
\\[-3ex]
\begin{subfigure}[b]{0.24\textwidth}
    \begin{tikzpicture}
\begin{axis}[
    width=1.25*\linewidth,
    height=\linewidth,
    title={Diving48, LP},
    title style={yshift=-8pt, font=\small},
    xlabel={Throughput (videos/s)},
    ylabel={Accuracy (\%)},
    xlabel style={yshift=5pt, font=\scriptsize},
    ylabel style={yshift=-5pt, font=\scriptsize},
    tick label style={font=\scriptsize},
    xmin=0, xmax=1130,
    ymin=19, ymax=43,
    grid=major,
    mark size=1.5pt,
    line width=1.2pt,
]

\addplot[color=lwpurple, mark=*, mark options={solid}]
coordinates {
    (1061.94, 27.77)
    (740.2, 33.3)
    (546.15, 35.23)
    (259.1, 36.8)
    (226.89, 39.49)
    (198.57, 39.8)
    (156.11, 39.85)
};

\addplot[color=tldrteal, mark=square*, mark options={solid}]
coordinates {
    (163.82, 22.54)
    (103.13, 24.77)
    (99.28, 26.55)
    (83.45, 26.6)
    (85.52, 27.01)
};

\addplot[color=rltamber, mark=triangle*, mark options={solid}]
coordinates {
    (105.25, 26.4)
    (105.25, 26.95)
    (105.25, 26.19)
};

\node[
    star,
    star points=5,
    star point ratio=2,
    draw=black,
    fill=white,
    inner sep=1pt,
] at (axis cs:105.25, 27.41) {};

\end{axis}
\end{tikzpicture}
    \phantomcaption
    \label{fig:ft:diving}
\end{subfigure}
\hfill
\begin{subfigure}[b]{0.24\textwidth}
    \begin{tikzpicture}
\begin{axis}[
    width=1.25*\linewidth,
    height=\linewidth,
    title={EK-100, LP},
    title style={yshift=-8pt, font=\small},
    xlabel={Throughput (videos/s)},
    ylabel={Accuracy (\%)},
    xlabel style={yshift=5pt, font=\scriptsize},
    ylabel style={yshift=-5pt, font=\scriptsize},
    tick label style={font=\scriptsize},
    xmin=0, xmax=1130,
    ymin=17, ymax=31,
    grid=major,
    mark size=1.5pt,
    line width=1.2pt,
]

\addplot[color=lwpurple, mark=*, mark options={solid}]
coordinates {
    (1061.94, 20.78)
    (740.2, 24.92)
    (546.15, 27.0)
    (259.1, 28.4)
    (226.89, 29.2)
    (198.57, 29.35)
    (156.11, 29.79)
};

\addplot[color=tldrteal, mark=square*, mark options={solid}]
coordinates {
    (163.82, 21.42)
    (155.15, 21.76)
    (99.28, 23.21)
    (68.4, 23.97)
    (85.52, 24.12)
};

\addplot[color=rltamber, mark=triangle*, mark options={solid}]
coordinates {
    (105.25, 18.81)
    (105.25, 19.14)
    (105.25, 19.02)
};

\node[
    star,
    star points=5,
    star point ratio=2,
    draw=black,
    fill=white,
    inner sep=1pt,
] at (axis cs:105.25, 24.12) {};

\end{axis}
\end{tikzpicture}
    \phantomcaption
    \label{fig:lp:diving}
\end{subfigure}
\hfill
\begin{subfigure}[b]{0.24\textwidth}
    \begin{tikzpicture}
\begin{axis}[
    width=1.25*\linewidth,
    height=\linewidth,
    title={Diving48, FT},
    title style={yshift=-8pt, font=\small},
    xlabel={Throughput (videos/s)},
    ylabel={Accuracy (\%)},
    xlabel style={yshift=5pt, font=\scriptsize},
    ylabel style={yshift=-5pt, font=\scriptsize},
    tick label style={font=\scriptsize},
    xmin=0, xmax=1130,
    ymin=61, ymax=94,
    grid=major,
    mark size=1.5pt,
    line width=1.2pt,
]

\addplot[color=lwpurple, mark=*, mark options={solid}]
coordinates {
    (1061.94, 81.37)
    (740.2, 86.04)
    (546.15, 88.48)
    (259.1, 90.0)
    (226.89, 89.34)
    (198.57, 89.49)
    (156.11, 89.7)
};

\addplot[color=tldrteal, mark=square*, mark options={solid}]
coordinates {
    (163.82, 64.52)
    (123.49, 73.6)
    (92.56, 76.65)
    (82.39, 76.85)
    (73.83, 77.16)
};

\addplot[color=rltamber, mark=triangle*, mark options={solid}]
coordinates {
    (105.25, 85.99)
    (105.25, 85.79)
    (105.25, 85.89)
};

\node[
    star,
    star points=5,
    star point ratio=2,
    draw=black,
    fill=white,
    inner sep=1pt,
] at (axis cs:105.25, 85.79) {};

\end{axis}
\end{tikzpicture}
    \phantomcaption
    \label{fig:ft:ek}
\end{subfigure}
\hfill
\begin{subfigure}[b]{0.24\textwidth}
    \begin{tikzpicture}
\begin{axis}[
    width=1.25*\linewidth,
    height=\linewidth,
    title={EK-100, FT},
    title style={yshift=-8pt, font=\small},
    xlabel={Throughput (videos/s)},
    ylabel={Accuracy (\%)},
    xlabel style={yshift=5pt, font=\scriptsize},
    ylabel style={yshift=-5pt, font=\scriptsize},
    tick label style={font=\scriptsize},
    xmin=0, xmax=1130,
    ymin=37, ymax=55,
    grid=major,
    mark size=1.5pt,
    line width=1.2pt,
]

\addplot[color=lwpurple, mark=*, mark options={solid}]
coordinates {
    (1061.94, 39.24)
    (740.2, 44.63)
    (546.15, 46.84)
    (259.1, 49.95)
    (226.89, 51.8)
    (198.57, 52.1)
    (156.11, 52.83)
};

\addplot[color=tldrteal, mark=square*, mark options={solid}]
coordinates {
    (163.82, 45.64)
    (103.13, 49.62)
    (79.61, 51.74)
    (83.45, 52.27)
    (87.06, 52.7)
};

\addplot[color=rltamber, mark=triangle*, mark options={solid}]
coordinates {
    (105.25, 50.93)
    (105.25, 51.12)
    (105.25, 51.25)
};

\node[
    star,
    star points=5,
    star point ratio=2,
    draw=black,
    fill=white,
    inner sep=1pt,
] at (axis cs:105.25, 52.99) {};

\end{axis}
\end{tikzpicture}
    \phantomcaption
    \label{fig:lp:ek}
\end{subfigure}
\\[-3ex]
\begin{subfigure}[b]{0.24\textwidth}
    \begin{tikzpicture}
\begin{axis}[
    width=1.25*\linewidth,
    height=\linewidth,
    title={Jester, LP},
    title style={yshift=-8pt, font=\small},
    xlabel={Throughput (videos/s)},
    ylabel={Accuracy (\%)},
    xlabel style={yshift=5pt, font=\scriptsize},
    ylabel style={yshift=-5pt, font=\scriptsize},
    tick label style={font=\scriptsize},
    xmin=0, xmax=1130,
    ymin=61, ymax=90,
    grid=major,
    mark size=1.5pt,
    line width=1.2pt,
]

\addplot[color=lwpurple, mark=*, mark options={solid}]
coordinates {
    (1061.94, 81.21)
    (740.2, 84.04)
    (546.15, 84.64)
    (259.1, 85.21)
    (226.89, 85.32)
    (198.57, 85.62)
    (156.11, 85.39)
};

\addplot[color=tldrteal, mark=square*, mark options={solid}]
coordinates {
    (163.82, 65.29)
    (103.13, 71.87)
    (79.61, 74.52)
    (68.4, 75.17)
    (62.36, 75.42)
};

\addplot[color=rltamber, mark=triangle*, mark options={solid}]
coordinates {
    (105.25, 66.64)
    (105.25, 66.17)
    (105.25, 64.75)
};

\node[
    star,
    star points=5,
    star point ratio=2,
    draw=black,
    fill=white,
    inner sep=1pt,
] at (axis cs:105.25, 74.61) {};

\end{axis}
\end{tikzpicture}
    \phantomcaption
    \label{fig:ft:jester}
\end{subfigure}
\hfill
\begin{subfigure}[b]{0.24\textwidth}
    \begin{tikzpicture}
\begin{axis}[
    width=1.25*\linewidth,
    height=\linewidth,
    title={Charades, LP},
    title style={yshift=-8pt, font=\small},
    xlabel={Throughput (videos/s)},
    ylabel={Accuracy (\%)},
    xlabel style={yshift=5pt, font=\scriptsize},
    ylabel style={yshift=-5pt, font=\scriptsize},
    tick label style={font=\scriptsize},
    xmin=0, xmax=1130,
    ymin=24, ymax=38,
    grid=major,
    mark size=1.5pt,
    line width=1.2pt,
]

\addplot[color=lwpurple, mark=*, mark options={solid}]
coordinates {
    (1061.94, 27.63)
    (740.2, 31.33)
    (546.15, 33.07)
    (259.1, 34.36)
    (226.89, 35.53)
    (198.57, 35.66)
    (156.11, 35.82)
};

\addplot[color=tldrteal, mark=square*, mark options={solid}]
coordinates {
    (163.82, 27.41)
    (103.13, 30.16)
    (79.61, 30.82)
    (68.4, 30.8)
    (62.36, 30.69)
};

\addplot[color=rltamber, mark=triangle*, mark options={solid}]
coordinates {
    (105.25, 25.51)
    (105.25, 25.53)
    (105.25, 25.4)
};

\node[
    star,
    star points=5,
    star point ratio=2,
    draw=black,
    fill=white,
    inner sep=1pt,
] at (axis cs:105.25, 30.35) {};

\end{axis}
\end{tikzpicture}
    \phantomcaption
    \label{fig:lp:jester}
\end{subfigure}
\hfill
\begin{subfigure}[b]{0.24\textwidth}
    \begin{tikzpicture}
\begin{axis}[
    width=1.25*\linewidth,
    height=\linewidth,
    title={Jester, FT},
    title style={yshift=-8pt, font=\small},
    xlabel={Throughput (videos/s)},
    ylabel={Accuracy (\%)},
    xlabel style={yshift=5pt, font=\scriptsize},
    ylabel style={yshift=-5pt, font=\scriptsize},
    tick label style={font=\scriptsize},
    xmin=0, xmax=1130,
    ymin=92, ymax=98,
    grid=major,
    mark size=1.5pt,
    line width=1.2pt,
]

\addplot[color=lwpurple, mark=*, mark options={solid}]
coordinates {
    (1061.94, 95.91)
    (740.2, 96.67)
    (546.15, 96.83)
    (259.1, 96.98)
    (226.89, 97.04)
    (198.57, 97.04)
    (156.11, 97.03)
};

\addplot[color=tldrteal, mark=square*, mark options={solid}]
coordinates {
    (163.82, 92.59)
    (103.13, 95.54)
    (79.61, 96.23)
    (82.39, 96.38)
    (62.36, 96.42)
};

\addplot[color=rltamber, mark=triangle*, mark options={solid}]
coordinates {
    (105.25, 96.23)
    (105.25, 96.42)
    (105.25, 96.44)
};

\node[
    star,
    star points=5,
    star point ratio=2,
    draw=black,
    fill=white,
    inner sep=1pt,
] at (axis cs:105.25, 96.72) {};

\end{axis}
\end{tikzpicture}
    \phantomcaption
    \label{fig:ft:charades}
\end{subfigure}
\hfill
\begin{subfigure}[b]{0.24\textwidth}
    \begin{tikzpicture}
\begin{axis}[
    width=1.25*\linewidth,
    height=\linewidth,
    title={Charades, FT},
    title style={yshift=-8pt, font=\small},
    xlabel={Throughput (videos/s)},
    ylabel={Accuracy (\%)},
    xlabel style={yshift=5pt, font=\scriptsize},
    ylabel style={yshift=-5pt, font=\scriptsize},
    tick label style={font=\scriptsize},
    xmin=0, xmax=1130,
    ymin=32, ymax=50,
    grid=major,
    mark size=1.5pt,
    line width=1.2pt,
]

\addplot[color=lwpurple, mark=*, mark options={solid}]
coordinates {
    (1061.94, 34.42)
    (740.2, 39.77)
    (546.15, 42.2)
    (259.1, 44.8)
    (226.89, 46.47)
    (198.57, 46.77)
    (156.11, 47.17)
};

\addplot[color=tldrteal, mark=square*, mark options={solid}]
coordinates {
    (163.82, 37.7)
    (103.13, 42.27)
    (79.61, 44.26)
    (68.4, 44.95)
    (62.36, 45.09)
};

\addplot[color=rltamber, mark=triangle*, mark options={solid}]
coordinates {
    (105.25, 44.2)
    (105.25, 44.45)
    (105.25, 44.62)
};

\node[
    star,
    star points=5,
    star point ratio=2,
    draw=black,
    fill=white,
    inner sep=1pt,
] at (axis cs:105.25, 45.11) {};

\end{axis}
\end{tikzpicture}
    \phantomcaption
    \label{fig:lp:charades}
\end{subfigure}
\\[-3ex]
\caption{
\textbf{Throughput (videos/s) at inference time.}
All measurements are taken on an NVIDIA L40S GPU with batch size 32 and \texttt{bfloat16} automatic mixed precision.
RLT models with different sparsity have the same throughput because RLT requires token masking through attention masking (not token dropping!) for batch sizes greater than 1.
Markers: LookWhen (\textcolor{lwpurple}{\raisebox{-0.3pt}{\Large$\bullet$}}), IV2 ($\star$), IV2+vid-TLDR (\textcolor{tldrteal}{$\blacksquare$}), and IV2+RLT (\textcolor{rltamber}{$\blacktriangle$}).
}
\label{fig:lp_ft_efficiency_speed}
\end{figure*}
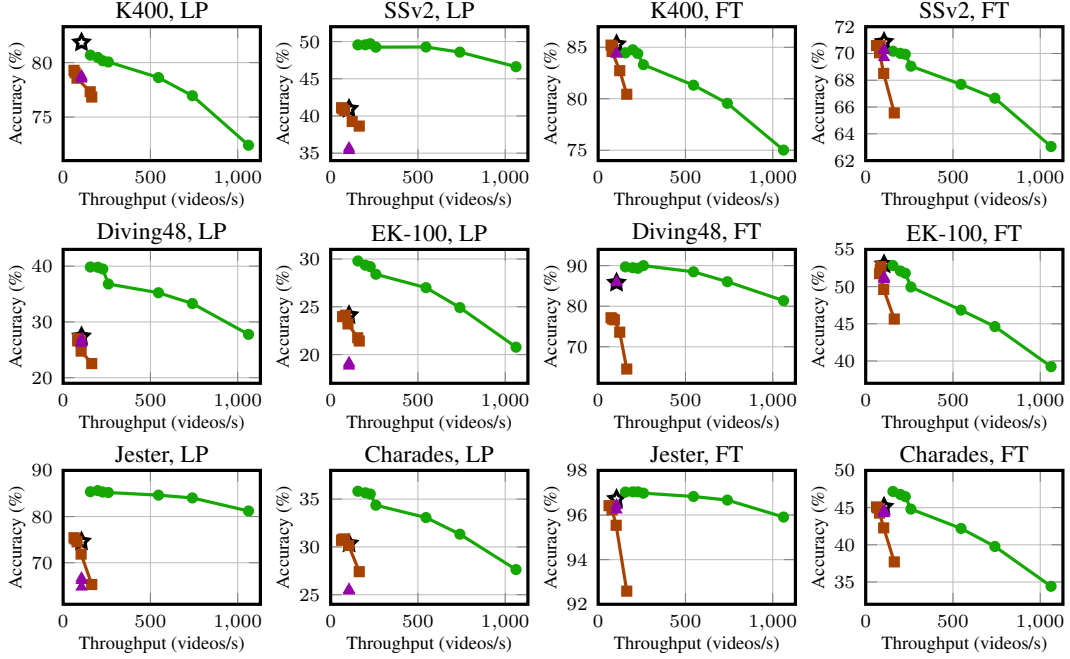

\begin{table}[ht!]
\centering
\setlength{\tabcolsep}{6pt}
\captionof{table}{
\textbf{Peak memory use (GB) at inference time.}
For RLT, all sparsity levels reach equal peak memory because RLT requires masking for batch processing; masking still computes all tokens, it just ignores their contribution.
For vid-TLDR, all sparsity levels are equal because the first layer processes \emph{all tokens} regardless of the token-merging schedule; it consumes much more memory because the token-merging layers are incompatible with Flash Attention \cite{dao2023flashattention2}.
All measurements are taken on an NVIDIA L40S GPU with batch size 32 and \texttt{bfloat16} automatic mixed precision.
}
\vspace{0.2cm}
\label{tab:datasets}
\begin{tabular}{cccccccccc}
\multicolumn{7}{c}{LookWhen (sparsity)} & & & \\
\cmidrule(lr){1-7}
95\% &
90\% &
85\% &
70\% &
65\% &
60\% &
50\% &
IV2 \cite{internvideo2} &
+RLT \cite{rlt} &
+vid-TLDR \cite{vid_tldr}
\\
\midrule
1.51&
1.51&
1.51&
1.68&
1.80&
1.91&
2.15&
2.76&
2.76&
16.59

\end{tabular}
\end{table}

\clearpage

\subsection{Learned Selection Examples}\label{app:selection_examples}

\def\cell{0.105\textwidth}

\captionof{figure}{Example from Kinetics-400.}
\setlength{\tabcolsep}{1.5pt}
\begin{tabular}{c c c c c c c c c}
\centering

\raisebox{16pt}{\makecell{frames\\1-8}}
& \includegraphics[width=\cell]{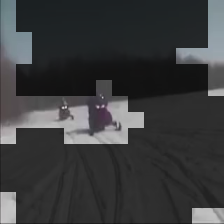}
& \includegraphics[width=\cell]{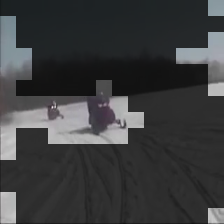}
& \includegraphics[width=\cell]{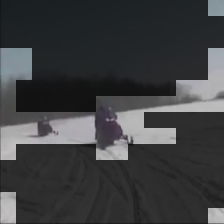}
& \includegraphics[width=\cell]{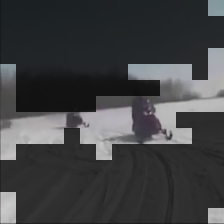}
& \includegraphics[width=\cell]{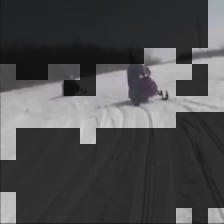}
& \includegraphics[width=\cell]{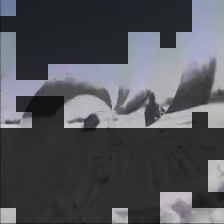}
& \includegraphics[width=\cell]{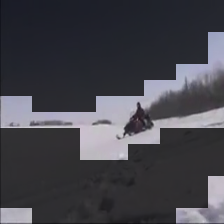}
& \includegraphics[width=\cell]{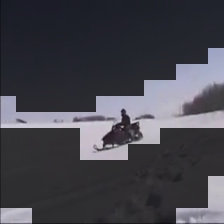} \\

\raisebox{16pt}{\makecell{frames\\9-16}}
& \includegraphics[width=\cell]{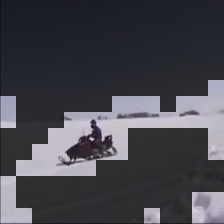}
& \includegraphics[width=\cell]{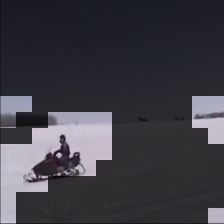}
& \includegraphics[width=\cell]{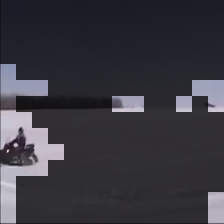}
& \includegraphics[width=\cell]{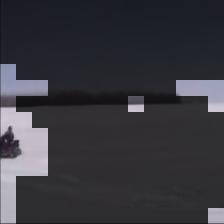}
& \includegraphics[width=\cell]{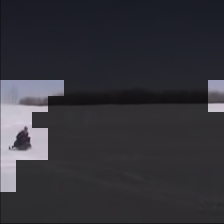}
& \includegraphics[width=\cell]{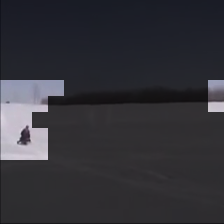}
& \includegraphics[width=\cell]{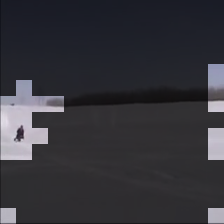}
& \includegraphics[width=\cell]{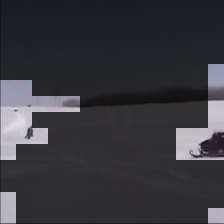} \\
\end{tabular}
\def\cell{0.105\textwidth}

\setlength{\tabcolsep}{1.5pt}
\captionof{figure}{Example from Something-Something-v2.}
\begin{tabular}{c c c c c c c c c}
\centering

\raisebox{16pt}{\makecell{frames\\1-8}}
& \includegraphics[width=\cell]{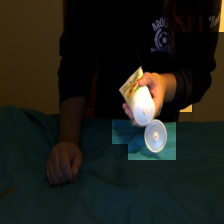}
& \includegraphics[width=\cell]{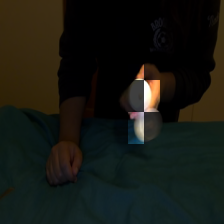}
& \includegraphics[width=\cell]{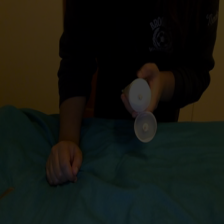}
& \includegraphics[width=\cell]{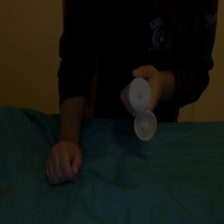}
& \includegraphics[width=\cell]{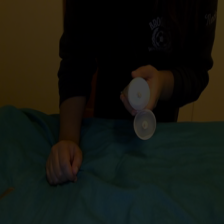}
& \includegraphics[width=\cell]{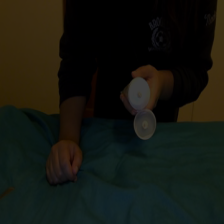}
& \includegraphics[width=\cell]{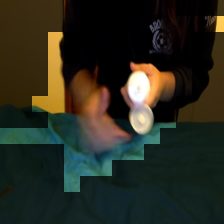}
& \includegraphics[width=\cell]{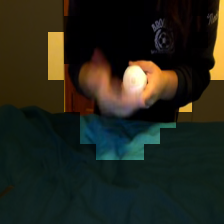} \\

\raisebox{16pt}{\makecell{frames\\9-16}}
& \includegraphics[width=\cell]{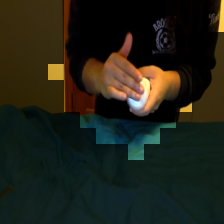}
& \includegraphics[width=\cell]{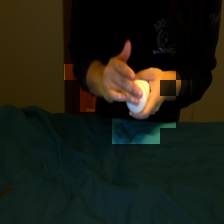}
& \includegraphics[width=\cell]{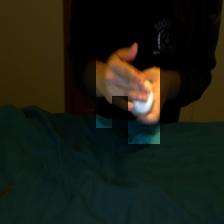}
& \includegraphics[width=\cell]{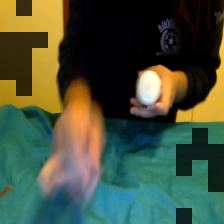}
& \includegraphics[width=\cell]{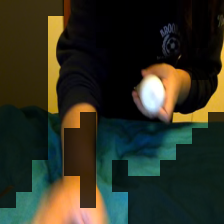}
& \includegraphics[width=\cell]{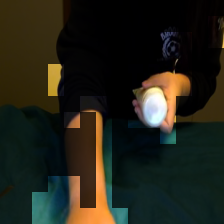}
& \includegraphics[width=\cell]{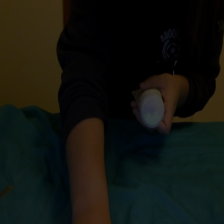}
& \includegraphics[width=\cell]{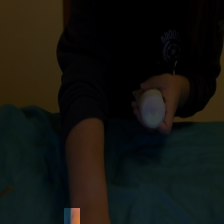} \\

\end{tabular}
\def\cell{0.105\textwidth}

\setlength{\tabcolsep}{1.5pt}
\captionof{figure}{Example from Epic-Kitchens-100.}
\begin{tabular}{c c c c c c c c c}
\centering

\raisebox{16pt}{\makecell{frames\\1-8}}
& \includegraphics[width=\cell]{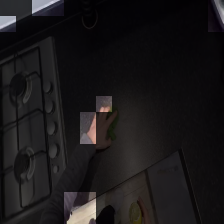}
& \includegraphics[width=\cell]{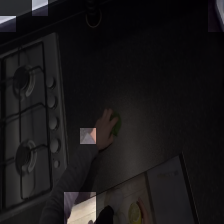}
& \includegraphics[width=\cell]{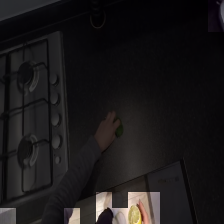}
& \includegraphics[width=\cell]{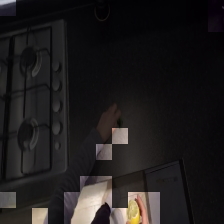}
& \includegraphics[width=\cell]{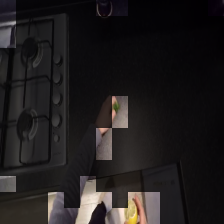}
& \includegraphics[width=\cell]{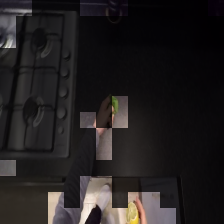}
& \includegraphics[width=\cell]{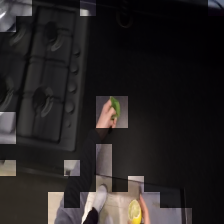}
& \includegraphics[width=\cell]{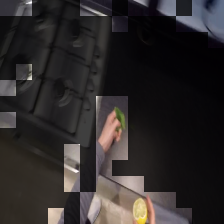} \\

\raisebox{16pt}{\makecell{frames\\9-16}}
& \includegraphics[width=\cell]{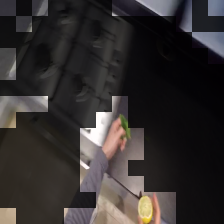}
& \includegraphics[width=\cell]{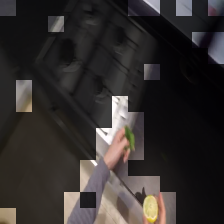}
& \includegraphics[width=\cell]{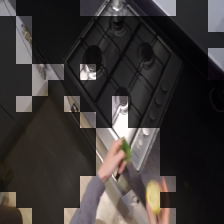}
& \includegraphics[width=\cell]{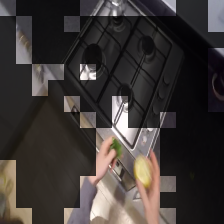}
& \includegraphics[width=\cell]{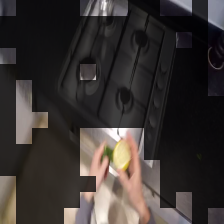}
& \includegraphics[width=\cell]{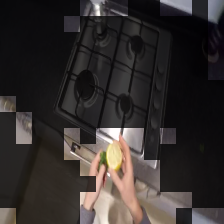}
& \includegraphics[width=\cell]{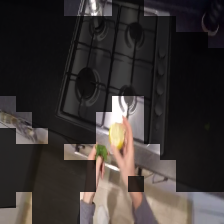}
& \includegraphics[width=\cell]{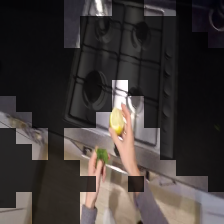} \\

\end{tabular}
\clearpage

\def\cell{0.105\textwidth}

\setlength{\tabcolsep}{1.5pt}
\captionof{figure}{Example from Diving48.}
\begin{tabular}{c c c c c c c c c}
\centering

\raisebox{16pt}{\makecell{frames\\1-8}}
& \includegraphics[width=\cell]{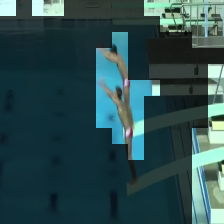}
& \includegraphics[width=\cell]{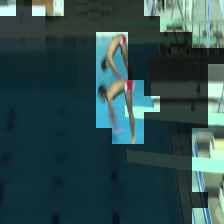}
& \includegraphics[width=\cell]{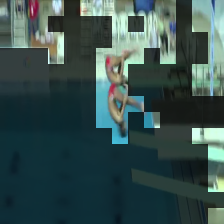}
& \includegraphics[width=\cell]{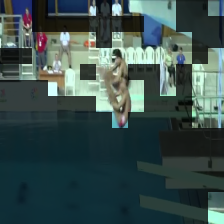}
& \includegraphics[width=\cell]{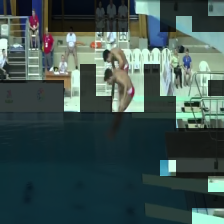}
& \includegraphics[width=\cell]{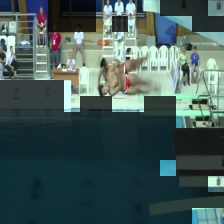}
& \includegraphics[width=\cell]{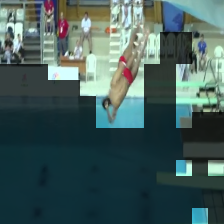}
& \includegraphics[width=\cell]{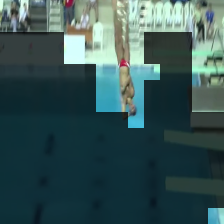} \\

\raisebox{16pt}{\makecell{frames\\9-16}}
& \includegraphics[width=\cell]{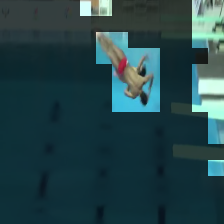}
& \includegraphics[width=\cell]{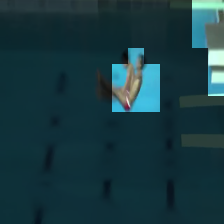}
& \includegraphics[width=\cell]{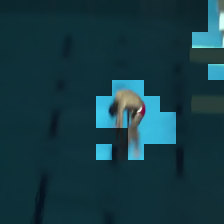}
& \includegraphics[width=\cell]{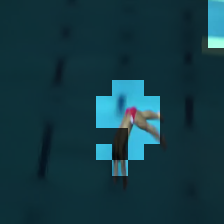}
& \includegraphics[width=\cell]{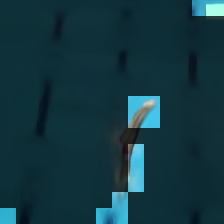}
& \includegraphics[width=\cell]{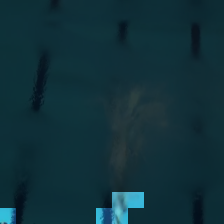}
& \includegraphics[width=\cell]{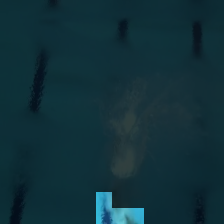}
& \includegraphics[width=\cell]{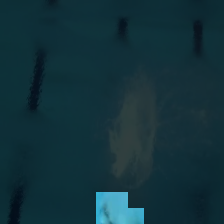} \\

\end{tabular}
\def\cell{0.105\textwidth}

\setlength{\tabcolsep}{1.5pt}
\captionof{figure}{Example from Jester.}
\begin{tabular}{c c c c c c c c c}
\centering

\raisebox{16pt}{\makecell{frames\\1-8}}
& \includegraphics[width=\cell]{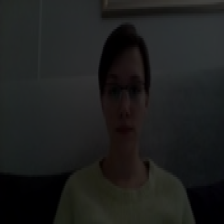}
& \includegraphics[width=\cell]{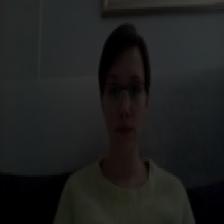}
& \includegraphics[width=\cell]{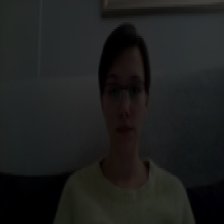}
& \includegraphics[width=\cell]{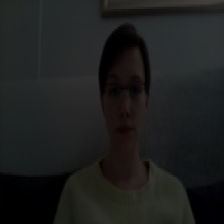}
& \includegraphics[width=\cell]{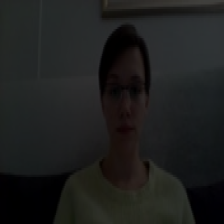}
& \includegraphics[width=\cell]{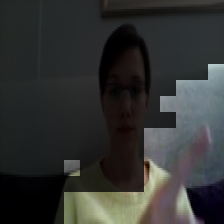}
& \includegraphics[width=\cell]{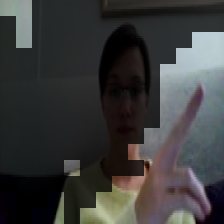}
& \includegraphics[width=\cell]{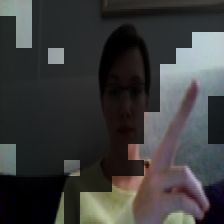} \\

\raisebox{16pt}{\makecell{frames\\9-16}}
& \includegraphics[width=\cell]{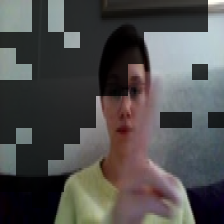}
& \includegraphics[width=\cell]{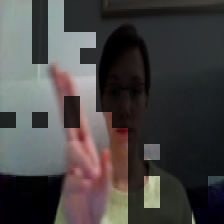}
& \includegraphics[width=\cell]{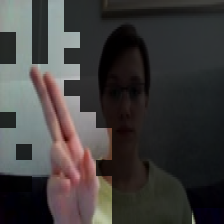}
& \includegraphics[width=\cell]{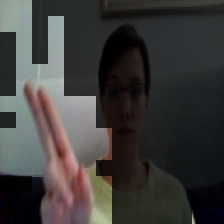}
& \includegraphics[width=\cell]{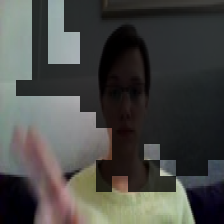}
& \includegraphics[width=\cell]{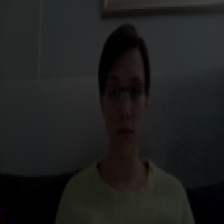}
& \includegraphics[width=\cell]{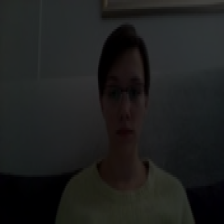}
& \includegraphics[width=\cell]{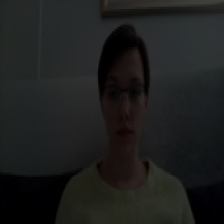} \\

\end{tabular}
\def\cell{0.105\textwidth}

\setlength{\tabcolsep}{1.5pt}
\captionof{figure}{Example from Charades.}
\begin{tabular}{c c c c c c c c c}
\centering

\raisebox{16pt}{\makecell{frames\\1-8}}
& \includegraphics[width=\cell]{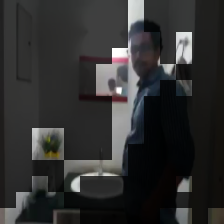}
& \includegraphics[width=\cell]{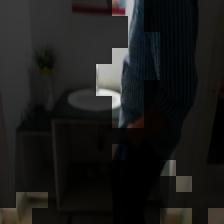}
& \includegraphics[width=\cell]{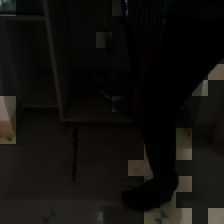}
& \includegraphics[width=\cell]{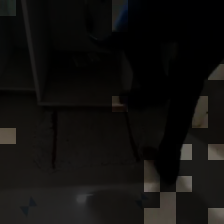}
& \includegraphics[width=\cell]{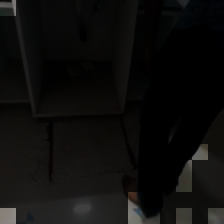}
& \includegraphics[width=\cell]{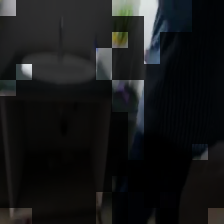}
& \includegraphics[width=\cell]{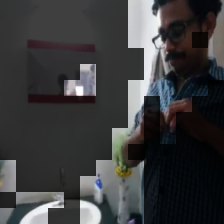}
& \includegraphics[width=\cell]{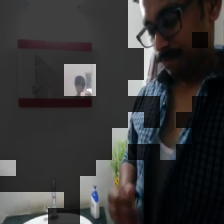} \\

\raisebox{16pt}{\makecell{frames\\9-16}}
& \includegraphics[width=\cell]{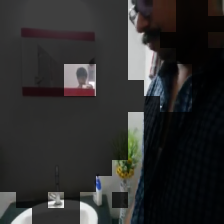}
& \includegraphics[width=\cell]{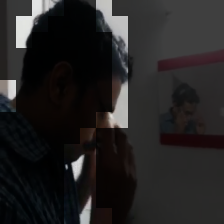}
& \includegraphics[width=\cell]{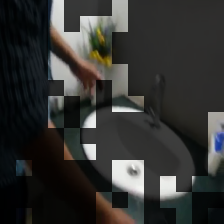}
& \includegraphics[width=\cell]{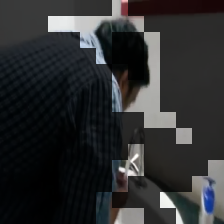}
& \includegraphics[width=\cell]{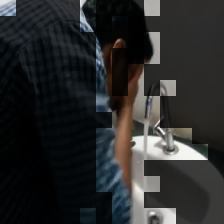}
& \includegraphics[width=\cell]{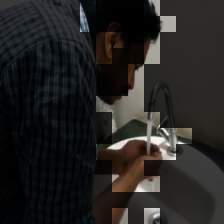}
& \includegraphics[width=\cell]{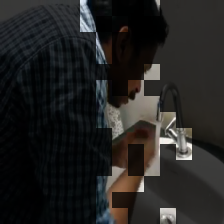}
& \includegraphics[width=\cell]{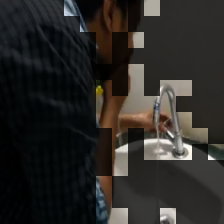} \\

\end{tabular}
\clearpage

\subsection{Implementation Details}
\label{appendix:implementation}

\subsubsection{Pre-training Settings}
We follow the data augmentation and input size used in fine-tuning (Tab.~\ref{tab:ft_config}).
We pre-train for 20 epochs but each video is actually seen 20$\times$8 times because of batch repetition, which reduces the fraction of time spent data loading and teacher processing.
Specifically, we load a batch, compute DINOv3 representations, then horizontally flip and stack both videos and DINOv3 representations (to align them); this is the first repeat.
We repeat another 4$\times$ by mixing up \cite{zhang2017mixup} videos and their corresponding DINOv3 representations.
We thus compute the teacher once for every 8 times the selector-extractor is computed (which is more efficient because it is sparse).
For ablations, we pre-train for 10 epochs.
We did not tune the pre-training learning rate, we used a max value of 0.0001 and a batch size of 16 (repeated 8$\times$ per above).

\subsubsection{Fine-tuning Settings}

For all datasets, we first train a linear probe for 100 epochs using the settings in Table~\ref{tab:lp_config}, and initialize the fine-tuning classification head from this probe.
For ablations, we subsample up to 20K training examples per dataset, uniformly across classes, and evaluate on the full validation set. For all other experiments, we use the full training data.
For the efficiency results in Fig.~\ref{fig:lp_ft_efficiency}, we follow the settings in Tab.~\ref{tab:ft_config}.
For Tab.~\ref{tab:super_table}, we use the same settings but train for 30 epochs on K400 and SSv2. All ablations use the same configuration with 20 training epochs.
Following VideoMAE~\citep{videomae}, we use dense sampling~\citep{dense1, dense2} on K400, and TSN-style sampling~\citep{tsn} on SSv2 and the remaining datasets.
\\
\\
Importantly, we run the exact same learning-rate sweep separately for IV2, IV2+RLT, and our LookWhen.
For IV2+vid-TLDR, we follow the original paper \cite{vid_tldr} by tuning the merging schedule.
Specifically, we first make 5 sparsity levels that merge 500, 700, 1000, 1400, or 1800 tokens.
For each sparsity level, we make 4 schedules (following templates in vid-TLDR's Table A) and choose the schedule that performs best.

\begin{table}[ht!]
\centering
\scriptsize
\setlength{\tabcolsep}{6.8pt}
\captionof{table}{
\textbf{Linear probing configuration.}
}
\label{tab:lp_config}
\begin{tabular}{ll}
\toprule
Config & Value \\
\midrule
Optimizer & AdamW \\
Base learning rate & 1e-4, 5e-4, 1e-3, 5e-3 \\
Batch size & 256 \\
Learning rate schedule & Cosine annealing \\
\bottomrule
\end{tabular}
\end{table}
\newcommand{\cmark}{\ding{51}}
\newcommand{\xmark}{\ding{55}}

\begin{table}[ht!]
\centering
\scriptsize
\setlength{\tabcolsep}{4.5pt}
\caption{
\textbf{Fine-tuning configuration.}
}
\label{tab:ft_config}
\begin{tabular}{lcccccc}
\toprule
Config & K400 & SSv2 & Diving48 & EK100 & Jester & Charades \\
\midrule
Optimizer                & \multicolumn{6}{c}{AdamW} \\
Weight decay             & \multicolumn{6}{c}{0.01} \\
Min.\ LR                 & \multicolumn{6}{c}{$5\times10^{-6}$} \\
Batch size               & \multicolumn{6}{c}{128} \\
LR schedule              & \multicolumn{6}{c}{Cosine annealing w/ linear warmup} \\
\midrule
Base LR (swept)          & \makecell{1e-5, 3e-5\\1e-4, 3e-4}
                         & \makecell{1e-4, 3e-4\\5e-4, 1e-3}
                         & 5e-4
                         & \makecell{1e-4, 3e-4\\5e-4, 1e-3}
                         & \makecell{1e-5, 3e-5\\1e-4, 3e-4}
                         & \makecell{1e-5, 3e-5\\1e-4, 3e-4} \\
Training epochs          & 10 & 10 & 50 & 10 & 10 & 50 \\
Warmup epochs            & 1  & 1  & 5  & 1  & 1  & 5  \\
\midrule
Input resolution         & \multicolumn{6}{c}{$224 \times 224$} \\
\# input frames          & \multicolumn{6}{c}{16} \\
Drop path rate           & \multicolumn{6}{c}{0.1} \\
\midrule
Random crop scale        & \multicolumn{6}{c}{$[0.5,\,1.0]$} \\
Horizontal flip          & \cmark & \xmark & \cmark & \xmark & \xmark & \xmark \\
Color jitter             & \multicolumn{6}{c}{\ ($p{=}0.8$,\ $b/c/s{=}0.4$,\ $h{=}0.1$)} \\
Gaussian noise           & \multicolumn{6}{c}{\ ($p{=}0.5$,\ $\sigma{=}0.1$)} \\
Batch renorm.\ aug.      & \multicolumn{6}{c}{\ ($p{=}0.5$)} \\
Mixup                    & \multicolumn{6}{c}{($\alpha{=}0.8$)} \\
CutMix                   & \multicolumn{6}{c}{($\alpha{=}1.0$)} \\
Label smoothing          & 0.1 & 0.1 & 0.1 & 0.1 & 0.1 & 0.0 \\
\bottomrule
\end{tabular}
\end{table}
\clearpage

\subsubsection{More results and Ablations}\label{appendix:more_ablations}

\begin{figure*}[h]
\centering
\begin{subfigure}[b]{0.24\textwidth}
    \begin{tikzpicture}
\begin{axis}[
    width=1.25*\linewidth,
    height=\linewidth,
    title={K400},
    title style={yshift=-8pt, font=\small},
    xlabel={Cumulative FLOPs (G$\times 10^8$)},
    ylabel={Accuracy (\%)},
    xlabel style={yshift=5pt, font=\scriptsize},
    ylabel style={yshift=-5pt, font=\scriptsize},
    tick label style={font=\scriptsize},
    xmin=0, xmax=38,
    ymin=78, ymax=86,
    grid=major,
    mark size=1.5pt,
    line width=1.2pt,
]
\addplot[color=black, mark=square, mark options={solid}]
coordinates {
    (3.62, 80.71)
    (7.23, 82.50)
    (10.80, 82.97)
    (14.50, 83.86)
    (18.10, 83.96)
    (21.70, 84.65)
    (25.30, 84.73)
    (28.90, 84.92)
    (32.50, 85.29)
    (36.20, 85.29)
};
\addplot[color=when, mark=*, mark options={solid}]
coordinates {
    (1.56, 78.74)
    (3.12, 79.30)
    (4.67, 80.34)
    (6.23, 80.65)
    (7.79, 81.16)
    (9.35, 82.14)
    (10.90, 82.48)
    (12.50, 82.94)
    (14.00, 83.07)
    (15.60, 83.31)
};
\node[anchor=north east, font=\scriptsize] at (rel axis cs:1,0) {$\times 10^8$};
\end{axis}
\end{tikzpicture}
    \phantomcaption
    \label{fig:lp:k400}
\end{subfigure}
\hspace{1cm}
\begin{subfigure}[b]{0.24\textwidth}
    \begin{tikzpicture}
\begin{axis}[
    width=1.25*\linewidth,
    height=\linewidth,
    title={SSv2},
    title style={yshift=-8pt, font=\small},
    xlabel={Cumulative FLOPs (G$\times 10^8$)},
    ylabel={Accuracy (\%)},
    xlabel style={yshift=5pt, font=\scriptsize},
    ylabel style={yshift=-5pt, font=\scriptsize},
    tick label style={font=\scriptsize},
    xmin=0, xmax=27,
    ymin=55, ymax=72,
    grid=major,
    mark size=1.5pt,
    line width=1.2pt,
]
\addplot[color=black, mark=square, mark options={solid}]
coordinates {
    (2.55, 56.91)
    (5.09, 62.84)
    (7.64, 66.02)
    (10.2, 67.43)
    (12.7, 68.61)
    (15.3, 69.46)
    (17.8, 69.85)
    (20.4, 70.43)
    (22.9, 70.81)
    (25.5, 70.86)
};
\addplot[color=when, mark=*, mark options={solid}]
coordinates {
    (1.10, 58.96)
    (2.20, 63.29)
    (3.29, 65.03)
    (4.39, 65.73)
    (5.49, 66.92)
    (6.59, 67.80)
    (7.68, 68.11)
    (8.78, 68.71)
    (9.88, 68.82)
    (11.0, 69.05)
};
\node[anchor=north east, font=\scriptsize] at (rel axis cs:1,0) {$\times 10^8$};
\end{axis}
\end{tikzpicture}
    \phantomcaption
    \label{fig:ft:ssv2}
\end{subfigure}
\\
\begin{subfigure}[b]{0.24\textwidth}
    \begin{tikzpicture}
\begin{axis}[
    width=1.25*\linewidth,
    height=\linewidth,
    title={Diving48},
    title style={yshift=-8pt, font=\small},
    xlabel={Cumulative FLOPs (G$\times 10^8$)},
    ylabel={Accuracy (\%)},
    xlabel style={yshift=5pt, font=\scriptsize},
    ylabel style={yshift=-5pt, font=\scriptsize},
    tick label style={font=\scriptsize},
    xmin=-0.5, xmax=12,
    ymin=22, ymax=94,
    grid=major,
    mark size=1.5pt,
    line width=1.2pt,
]
\addplot[color=black, mark=square, mark options={solid}]
coordinates {
    (0.227, 26.29)
    (1.13, 53.86)
    (2.27, 74.72)
    (3.40, 77.72)
    (4.53, 81.83)
    (5.66, 78.53)
    (6.80, 84.92)
    (7.93, 83.71)
    (9.06, 84.62)
    (10.2, 85.48)
    (11.3, 85.33)
};
\addplot[color=when, mark=*, mark options={solid}]
coordinates {
    (0.0976, 38.98)
    (0.488, 56.55)
    (0.976, 79.54)
    (1.46, 85.69)
    (1.95, 86.45)
    (2.44, 85.33)
    (2.93, 84.87)
    (3.42, 87.06)
    (3.91, 88.88)
    (4.39, 89.34)
    (4.88, 89.14)
};
\node[anchor=north east, font=\scriptsize] at (rel axis cs:1,0) {$\times 10^8$};
\end{axis}
\end{tikzpicture}
    \phantomcaption
    \label{fig:lp:k400}
\end{subfigure}
\hfill
\begin{subfigure}[b]{0.24\textwidth}
    \begin{tikzpicture}
\begin{axis}[
    width=1.25*\linewidth,
    height=\linewidth,
    title={EK-100},
    title style={yshift=-8pt, font=\small},
    xlabel={Cumulative FLOPs (G$\times 10^8$)},
    ylabel={Accuracy (\%)},
    xlabel style={yshift=5pt, font=\scriptsize},
    ylabel style={yshift=-5pt, font=\scriptsize},
    tick label style={font=\scriptsize},
    xmin=-0.2, xmax=11,
    ymin=28, ymax=55,
    grid=major,
    mark size=1.5pt,
    line width=1.2pt,
]
\addplot[color=black, mark=square, mark options={solid}]
coordinates {
    (1.01, 30.20)
    (2.03, 39.31)
    (3.04, 43.38)
    (4.05, 45.90)
    (5.07, 47.43)
    (6.08, 49.91)
    (7.09, 51.83)
    (8.11, 52.15)
    (9.12, 52.66)
    (10.1, 52.99)
};
\addplot[color=when, mark=*, mark options={solid}]
coordinates {
    (0.437, 34.28)
    (0.874, 38.44)
    (1.31, 41.15)
    (1.75, 42.53)
    (2.18, 44.17)
    (2.62, 46.91)
    (3.06, 47.20)
    (3.49, 49.19)
    (3.93, 49.53)
    (4.37, 49.95)
};
\node[anchor=north east, font=\scriptsize] at (rel axis cs:1,0) {$\times 10^8$};
\end{axis}
\end{tikzpicture}
    \phantomcaption
    \label{fig:ft:ssv2}
\end{subfigure}
\hfill
\begin{subfigure}[b]{0.24\textwidth}
    \begin{tikzpicture}
\begin{axis}[
    width=1.25*\linewidth,
    height=\linewidth,
    title={Jester},
    title style={yshift=-8pt, font=\small},
    xlabel={Cumulative FLOPs (G$\times 10^8$)},
    ylabel={Accuracy (\%)},
    xlabel style={yshift=5pt, font=\scriptsize},
    ylabel style={yshift=-5pt, font=\scriptsize},
    tick label style={font=\scriptsize},
    xmin=-0.2, xmax=19,
    ymin=86, ymax=98,
    grid=major,
    mark size=1.5pt,
    line width=1.2pt,
]
\addplot[color=black, mark=square, mark options={solid}]
coordinates {
    (1.79, 86.93)
    (3.58, 93.00)
    (5.36, 94.23)
    (7.15, 95.14)
    (8.94, 95.36)
    (10.7, 95.69)
    (12.5, 96.30)
    (14.3, 96.23)
    (16.1, 96.69)
    (17.9, 96.72)
};
\addplot[color=when, mark=*, mark options={solid}]
coordinates {
    (0.770, 93.10)
    (1.54, 94.98)
    (2.31, 95.33)
    (3.08, 95.86)
    (3.85, 96.10)
    (4.62, 96.37)
    (5.39, 96.54)
    (6.16, 96.77)
    (6.93, 96.92)
    (7.70, 96.98)
};
\node[anchor=north east, font=\scriptsize] at (rel axis cs:1,0) {$\times 10^8$};
\end{axis}
\end{tikzpicture}
    \phantomcaption
    \label{fig:ft:ssv2}
\end{subfigure}
\hfill
\begin{subfigure}[b]{0.24\textwidth}
    \begin{tikzpicture}
\begin{axis}[
    width=1.25*\linewidth,
    height=\linewidth,
    title={Charades},
    title style={yshift=-8pt, font=\small},
    xlabel={Cumulative FLOPs (G$\times 10^8$)},
    ylabel={Accuracy (\%)},
    xlabel style={yshift=5pt, font=\scriptsize},
    ylabel style={yshift=-5pt, font=\scriptsize},
    tick label style={font=\scriptsize},
    xmin=-0.3, xmax=7,
    ymin=29, ymax=47,
    grid=major,
    mark size=1.5pt,
    line width=1.2pt,
]
\addplot[color=black, mark=square, mark options={solid}]
coordinates {
    (0.120, 30.60)
    (0.602, 36.98)
    (1.20, 41.15)
    (1.81, 43.15)
    (2.41, 44.01)
    (3.01, 44.46)
    (3.61, 44.84)
    (4.21, 44.87)
    (4.82, 44.91)
    (5.42, 45.06)
    (6.02, 45.11)
};
\addplot[color=when, mark=*, mark options={solid}]
coordinates {
    (0.0519, 34.48)
    (0.259, 37.39)
    (0.519, 40.40)
    (0.778, 41.88)
    (1.04, 43.02)
    (1.30, 43.70)
    (1.56, 43.85)
    (1.82, 44.25)
    (2.08, 44.44)
    (2.34, 44.54)
    (2.59, 44.67)
};
\node[anchor=north east, font=\scriptsize] at (rel axis cs:1,0) {$\times 10^8$};
\end{axis}
\end{tikzpicture}
    \phantomcaption
    \label{fig:ft:ssv2}
\end{subfigure}
\\[-3ex]
\caption{
\textbf{Fine-tuning efficiency.}
We plot \emph{cumulative} fine-tuning cost vs. accuracy.
At 70\% sparsity, LookWhen (\textcolor{lwpurple}{\raisebox{-0.3pt}{\Large$\bullet$}}) reaches a given accuracy faster than the dense InternVideo2 ($\blacksquare$) during fine-tuning.
Each marker represents 1 epoch for EK-100 and Jester, and 5 epochs for Diving48 and Charades.
}
\label{fig:lp_ft_efficiency_training}
\end{figure*}
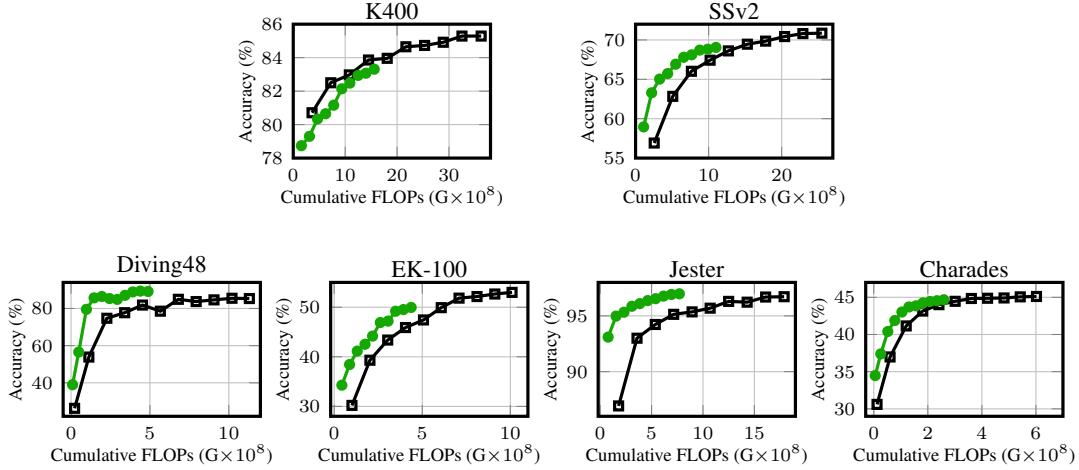

\begin{table*}[h]
  \centering
  \caption{
  \textbf{Full ablation over what to extract.}
  Settings are the same as the main-paper ablations.
  }
  \label{tab:extractor_video_target_full}
\centering
\scriptsize
\setlength{\tabcolsep}{5pt}

\begin{tabular}{ccccccccccccccccc}
& \multicolumn{4}{c}{DINOv3}
& \multicolumn{2}{c}{K400-20K}
& \multicolumn{2}{c}{SSv2-20K}
& \multicolumn{2}{c}{Diving48}
& \multicolumn{2}{c}{EK100-20K}
& \multicolumn{2}{c}{Jester-20K}
& \multicolumn{2}{c}{Charades}
\\
\cmidrule(lr){2-5}
\cmidrule(lr){6-7} \cmidrule(lr){8-9} \cmidrule(lr){10-11}  \cmidrule(lr){12-13}  \cmidrule(lr){14-15}  \cmidrule(lr){16-17}
IntVid2 & Vid & Frame & Patch & norm
& LP & FT
& LP & FT
& LP & FT
& LP & FT
& LP & FT
& LP & FT
\\
\midrule
\rowcolor{when!20} \ding{52} & \ding{52} & \ding{52} & \ding{52} & \ding{52} &
72.5 & 73.8 & 38.1 & \underline{50.1} & \underline{37.6} & 84.8 & \underline{13.5} & \underline{23.8} & \textbf{79.0} & \underline{95.2} & \textbf{31.3} & \textbf{39.3}\\
\ding{52} & \ding{52} & \ding{52} & \ding{56} & \ding{52} &
72.5 & 73.8 & \underline{38.6} & \underline{50.1} & 35.5 & 84.1 & \textbf{13.8} & \textbf{24.2} & 77.0 & \underline{95.2} & \underline{31.0} & \underline{39.1}\\
\ding{52} & \ding{52} & \ding{56} & \ding{52} & \ding{52} &
72.7 & 73.8 & \textbf{39.1} & 49.6 & 31.9 & 84.6 & \textbf{13.8} & \underline{23.8} & \underline{77.1} & \textbf{95.3} & 30.7 & 38.7\\
\ding{52} & \ding{52} & \ding{56} & \ding{56} & \ding{52} &
\textbf{73.3} & \textbf{74.0} & 37.9 & 48.0 & 29.9 & 81.9 & 12.8 & 23.3 & 69.0 & 94.6 & 30.3 & 37.4\\
\ding{52} & \ding{52} & \ding{56} & \ding{56} & \ding{56} &
72.1 & 73.3 & 31.5 & 45.1 & 32.1 & 84.2 & 11.3 & 22.4 & 63.9 & 94.9 & 29.5 & 36.9\\
\midrule
\ding{52} & \ding{56} & \ding{52} & \ding{52} & \ding{56} &
72.9 & 73.6 & 31.5 & 45.2 & 29.7 & 83.2 & 11.6 & 22.2 & 63.4 & 94.7 & 29.3 & 36.7\\
\ding{52} & \ding{56} & \ding{52} & \ding{56} & \ding{56} &
72.9 & 73.7 & 31.8 & 46.4 & 30.4 & 82.7 & 11.3 & 22.7 & 62.7 & 94.9 & 29.3 & 37.0\\
\ding{52} & \ding{56} & \ding{56} & \ding{52} & \ding{56} &
\underline{73.2} & \underline{73.9} & 32.0 & 45.0 & 29.4 & 83.2 & 11.3 & 21.6 & 60.8 & 94.4 & 28.9 & 36.2\\
\ding{52} & \ding{56} & \ding{56} & \ding{56} & \ding{56} &
\textbf{73.3} & \underline{73.9} & 32.1 & 45.6 & 27.9 & 83.1 & 11.0 & 22.0 & 61.6 & 94.4 & 29.0 & 36.3\\
\midrule
\ding{56} & \ding{52} & \ding{52} & \ding{52} & \ding{52} &
24.3 & 65.3 & 25.5 & \textbf{51.5} & 29.8 & 84.9 & 2.8 & 20.9 & 70.8 & \underline{95.2} & 14.9 & 35.5\\
\ding{56} & \ding{52} & \ding{52} & \ding{52} & \ding{56} &
63.8 & 68.2 & 21.8 & 43.4 & \textbf{37.8} & 84.7 & 6.6 & 21.2 & 56.3 & 94.9 & 23.0 & 34.3\\
\ding{56} & \ding{52} & \ding{56} & \ding{56} & \ding{52} &
17.9 & 61.1 & 26.7 & 48.8 & 26.7 & \textbf{85.7} & 2.6 & 18.7 & 70.4 & \underline{95.2} & 14.0 & 30.5\\
\ding{56} & \ding{56} & \ding{52} & \ding{52} & \ding{56} &
53.9 & 68.0 & 25.1 & 45.6 & 31.6 & \underline{85.6} & 6.5 & 21.0 & 77.0 & 94.9 & 21.6 & 36.0\\
\end{tabular}

\end{table*}

\begin{table}[h]
\centering
\scriptsize
\setlength{\tabcolsep}{7.1pt}
\captionof{table}{
\textbf{Ablating pre-training data.}
Pre-training on unlabeled K400 mostly helps downstream accuracy on K400, Diving48, and Charades; pre-training on unlabeled SSv2 mostly improves SSv2.
}
\label{tab:datasets}
\begin{tabular}{lrcccccccccccc}

&
& \multicolumn{2}{c}{K400-20K}
& \multicolumn{2}{c}{SSv2-20K}
& \multicolumn{2}{c}{Diving48}
& \multicolumn{2}{c}{EK100-20K}
& \multicolumn{2}{c}{Jester-20K}
& \multicolumn{2}{c}{Charades}
\\
\cmidrule(lr){3-4} \cmidrule(lr){5-6} \cmidrule(lr){7-8} \cmidrule(lr){9-10}  \cmidrule(lr){11-12}  \cmidrule(lr){13-14}
Dataset
& Steps
& LP & FT
& LP & FT
& LP & FT
& LP & FT
& LP & FT
& LP & FT
\\
\midrule
\rowcolor{when!20} K400+SSv2 & 1$\times$ &
\underline{72.9} & \textbf{73.6} & \underline{31.5} & \textbf{45.2} & \underline{29.7} & \textbf{83.2} & \textbf{11.6} & \textbf{22.2} & \underline{63.4} & \textbf{94.7} & \textbf{29.3} & \textbf{36.7}\\
K400 & 0.6$\times$ &
\textbf{73.0} & \underline{73.3} & 24.5 & 42.8 & \textbf{31.9} & \underline{83.1} & 9.8 & \underline{20.8} & \textbf{64.1} & \underline{94.6} & \underline{28.2} & \underline{36.0} \\
SSv2 & 0.4$\times$ &
55.2 & 63.3 & \textbf{32.1} & \underline{43.9} & 21.4 & 71.8 & \underline{10.2} & 20.2 & \underline{63.4} & 94.3 & 23.3 & 32.1\\

\end{tabular}
\end{table}



\end{document}